\pdfoutput=1

\documentclass[11pt]{article}

\usepackage[]{acl}

\usepackage{amsmath,amsfonts,bm}

\def\eqref#1{equation~\ref{#1}}

\def\1{\bm{1}}

\DeclareMathAlphabet{\mathsfit}{\encodingdefault}{\sfdefault}{m}{sl}
\SetMathAlphabet{\mathsfit}{bold}{\encodingdefault}{\sfdefault}{bx}{n}

\usepackage{times}
\usepackage{latexsym}
\usepackage[T1]{fontenc}

\usepackage[utf8]{inputenc}

\usepackage{microtype}
\usepackage{algorithm}
\usepackage{algorithmic}
\usepackage{comment}
\usepackage{todonotes}
\usepackage{color,soul}
\usepackage{graphicx}
\usepackage{pbox}
\usepackage{subcaption}
\usepackage{epstopdf}
\usepackage{xstring}
\usepackage{algorithm}
\usepackage{algorithmic}
\usepackage[normalem]{ulem}
\usepackage{booktabs}
\useunder{\uline}{\ul}{}
\usepackage{enumitem,kantlipsum}

\newcommand{\model}{\mathbb{M}}

\title{Analyzing Encoded Concepts in Transformer Language Models}

\author{Hassan Sajjad$^{\diamond}$ ~ Nadir Durrani$^{\diamond}$ ~ Fahim Dalvi$^{\diamond}$ ~ Firoj Alam$^{\diamond}$ \\\textbf{~ Abdul Rafae Khan$^{\dagger}$ ~ Jia Xu$^{\dagger}$} \\ 
{\tt \{hsajjad,ndurrani,faimaduddin,fialam\}@hbku.edu.qa} \\ 
\textsuperscript{$\diamond$}Qatar Computing Research Institute, HBKU Research Complex, Qatar \\\\ 
{\tt \{akhan4,jxu70\}@stevens.edu}\\
\textsuperscript{$\dagger$}School of Engineering and Science, Steven Institute of Technology, USA \\ }

\begin{document}
\maketitle
\begin{abstract}

We propose a novel framework \texttt{ConceptX}, to analyze how 
latent concepts 
are encoded in representations 
learned within pre-trained language models. 
It uses 
clustering to discover the encoded concepts and explains them by aligning with a large set of human-defined concepts. Our analysis on seven transformer language models reveal interesting insights: i) the latent space within the learned representations overlap with different linguistic concepts to a varying degree, ii) the lower layers in the model are dominated by lexical concepts (e.g., affixation), 
whereas the core-linguistic concepts (e.g., morphological or syntactic relations) are better represented in the middle and higher layers, iii) some encoded concepts are multi-faceted and cannot be adequately explained using the existing human-defined concepts.\footnote{The code is available at \url{https://github.com/hsajjad/ConceptX}.} 

\end{abstract}

\section{Introduction}
\label{sec:introduction}

Contextualized word representations learned in deep neural network models (DDNs) 
capture rich concepts making them ubiquitous for
transfer learning towards downstream NLP. Despite their revolution, the blackbox nature of the deep NLP models is a major 
bottle-neck for their large scale adaptability. Understanding the inner dynamics of these models is important to ensure fairness, robustness, reliability and control. 

A plethora of research has been carried out to probe DNNs for the linguistic knowledge (e.g. morphology, syntactic and semantic roles) captured within the learned representations. A commonly used framework to gauge how well linguistic information can be extracted
from these models is the \emph{Probing Framework} \cite{hupkes2018visualisation}, where they train an auxiliary classifier using representations as features to predict the property of interest. The performance of the classifier reflects the amount of knowledge learned within representations.
To this end, the researchers have analyzed what knowledge is learned within the representations through relevant extrinsic phenomenon varying from word morphology \cite{vylomova2016word,belinkov:2017:acl} 
to high level concepts such as syntactic structure \cite{blevins-etal-2018-deep,marvin-linzen-2018-targeted} and semantics \cite{qian-qiu-huang:2016:P16-11, NEURIPS2019_159c1ffe,belinkov:2017:ijcnlp} 
or more generic properties \cite{adi2016fine,rogers-etal-2020-primer}. 

In this work, we approach the 
representation analysis from a different angle and present a novel framework \texttt{ConceptX}. In contrast to relying on the prediction capacity of the representations, we analyze the latent concepts learned within these representations and how knowledge is structured, 
using an unsupervised method.
More specifically, we question: i) do the representations encode knowledge inline with linguistic properties such as word morphology and semantics? ii) which properties dominate the overall structure in these representations? iii) 
does the model learn any novel concepts beyond linguistic properties?  
Answers to these questions reveal how deep neural network models structure language information to learn a task.

\begin{figure*}[!t] 
\begin{center}
        \includegraphics[width=0.9\textwidth]{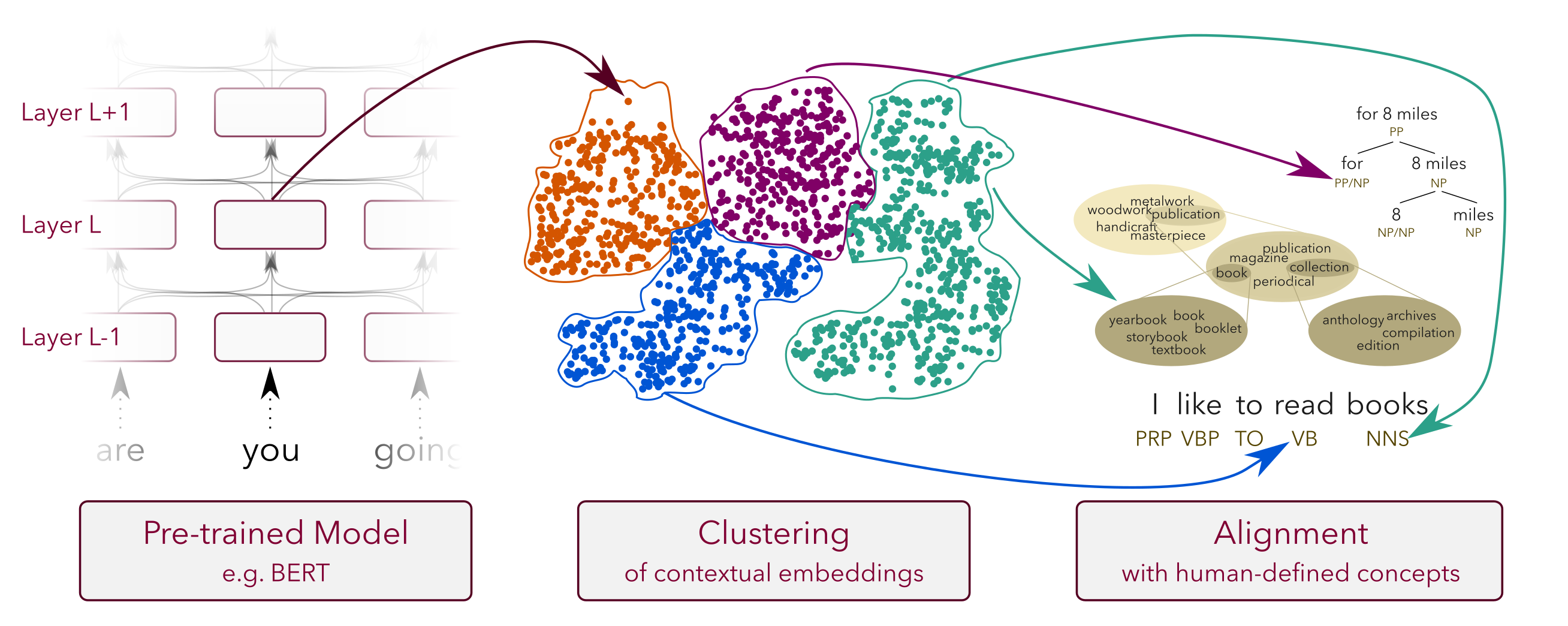}
      
      \caption{\textbf{ConceptX}: i) Extract representations from trained model, ii) Cluster the representations to obtain encoded concepts, iii) Align the concepts to human-defined concepts}
        \label{fig:encodedconcept}
        \end{center}
        
\end{figure*}

Our inspiration to use the term \textit{concept} comes from \textit{``concept based explanation''} in computer vision~\cite{kim2018interpretability,ghorbani2019towards,chen2020concept}. \newcite{stock2010concepts} defined a concept as ``a class containing certain objects as elements, where the objects have certain properties''. We define an \textit{encoded concept as a cluster of context-aware latent representations of words, where the representations are encoder layer outputs.}

%

Our framework clusters contextualized representations using agglomerative hierarchical clustering~\citep{gowda1978agglomerative}. The resulting clusters represent \emph{encoded concepts}, captured within the learned representations 
(Please see Figure \ref{fig:encodedconcept} 
 for illustration).
We then use a novel alignment function that measures the amount of overlap between \emph{encoded concepts} and a range of pre-defined 
categories (that we call as \emph{human-defined concepts} in this paper). We experimented with 
affixes, casing, 
morphological, syntactic, 
semantic, WordNet \cite{miller1995wordnet}, and psycholinguistic concepts (LIWC \newcite{pennebaker2001linguistic}). The use of such a diverse set of human-defined concepts 
enables us to cover various abstractions of language. 
In Figure 
\ref{fig:aligned_concept_examples} we present a few examples of human-defined concepts that were aligned with the encoded concepts.

We carry out our study on seven pre-trained transformer models such as BERT~\cite{devlin-etal-2019-bert} and XLM-RoBERTa \cite{xlm-roberta}, with varying optimization functions, architectural details and training data. Some notable findings emerging from our analysis are as follows:

\begin{itemize}
    \item Shallow concepts such as lexical ngrams or suffixes are predominantly captured in the lower layers of the network.
    \item 
    WordNet and psycholinguistic-based concepts (LIWC) are also learned in the lower layers.
    \item Middle and higher layers encode concepts that capture core linguistic properties such as morphology, semantics and syntax. 
    \item 
    Roughly 50\% of the encoded concepts adhere to our suite of human-defined linguistic concepts.
    \item The models learn novel concepts that are multi-faceted and cannot be adequately explained using the existing human-defined concepts.
\end{itemize}

\noindent Our contributions in this paper are as follow: i) We present 
\texttt{ConceptX}, a framework that interprets encoded concepts in the learned representation by 
measuring their alignment to the human-defined 
concepts. 
ii) We provide a qualitative and quantitative evidence of how knowledge is structured within deep NLP models with respect to a large suite of human-defined concepts. 


\section{Related Work}
\label{sec:related_work}

Most of the work done on interpretability in deep NLP addresses two questions in particular: {\em(i)} what linguistic (and non-linguistic) knowledge is learned within contextualized representations, \emph{Concept Analysis} and {\em(ii)} how  this  information  is  utilized  in  the  decision  making  process, \emph{Attribution Analysis} \cite{sajjad-etal-2021-fine}. The former thrives on post-hoc decomposability, where we analyze representations to uncover linguistic 
phenomenon that are captured as the network is trained towards any NLP task \cite{adi2016fine,conneau2018you,liu-etal-2019-linguistic,tenney-etal-2019-bert,belinkov-etal-2020-analysis} and the latter characterize the role of model components and input features towards a specific prediction \cite{linzen_tacl,gulordava-etal-2018-colorless,marvin-linzen-2018-targeted}. Our work falls into the former category. 

Previous studies 
have explored visualization methods to analyze the  learned representations \cite{karpathy2015visualizing,kadar2016representation}, attention heads \cite{clark-etal-2019-bert, vig-2019-multiscale}, language compositionality \cite{li-etal-2016-visualizing} etc. A more commonly used framework analyzes representations by correlating parts of the neural network with linguistic properties, 
by training a classifier to predict a feature of interest~\cite{adi2016fine,belinkov:2017:acl,conneau2018you}. Several researchers used probing classifiers for investigating the contextualized representations learned from a variety of neural language models on a variety of character- \cite{durrani-etal-2019-one}, word- \cite{liu-etal-2019-linguistic}  or 
sub-sentence level 
\cite{tenney-etal-2019-bert} linguistic tasks. Rather than analyzing the representations as a whole, several researchers also explored identifying salient neurons within the model that capture different properties \cite{dalvi:2019:AAAI, durrani-etal-2020-analyzing, suau2020finding, Mu-Nips} or are salient for the model irrespective of the property \cite{bau2018identifying, wu:2020:acl}.

Our work is inline with 
\cite{michael-etal-2020-asking, dalvi2022discovering}, who analyzed latent concepts learned in pre-trained models. 
\newcite{michael-etal-2020-asking} used a binary classification task to induce latent concepts relevant to a task and showed the presence of linguistically motivated and novel concepts in the representation. However, different from them, we 
analyze 
representations in an unsupervised fashion. 
\newcite{dalvi2022discovering} used human-in-the-loop to analyze latent spaces in BERT. Our framework uses human-defined concepts to automatically generate explanations for the latent concepts. This enabled us to scale our study to many transformer models. 

In a similar work, \newcite{mamou2020emergence} applied manifold analysis technique to understand the amount of information
stored about object categories per unit. 
Our approach does away from the methodological limitations of probing framework such as complexity of the probes, effect of randomness etc \cite{probingLimitations}. However, it is important to mention that the two frameworks are orthogonal and complement each other. 



\section{Methodology}
\label{sec:methodology}
A vector representation in the neural network model is composed of feature attributes of the input words. We group the encoded vector representations 
using 
a clustering approach discussed below. 
The underlying clusters, that we term as the \emph{encoded concepts}, are then matched with the human-defined concepts using an alignment function.
Formally, consider a Neural Network (NN) model $\model$ with $L$ encoder layers $\{l_1, l_2 , ...l_l, ... , l_L\}$, with $H$ hidden nodes per layer. An input sentence consisting of $M$ words $w_1, w_2, ...w_i, ... , w_M$ is fed into a NN. For 
each input word $i$, we compute the node output (after applying the activation functions) $y_h^l(w_i)$ of every hidden node $h\in\{1,...,H\}$ in each layer $l$, where $\overrightarrow{y}^l(w_i)$ is the vector representation composing the outputs of all hidden nodes in layer $l$ for $w_i$. Our goal is to cluster representations $\overrightarrow{y}^l$, from a large training data to obtain \emph{encoded concepts}. 
We then 
align these with various human-defined concepts to obtain an explanation of them to build an understanding of how these concepts are represented across 
the network.


\subsection{Clustering}
\label{sec:clustering}

We use agglomerative hierarchical clustering~\citep{gowda1978agglomerative}, which we found to be effective for this task. 
It
assigns each word to a separate cluster and then iteratively combines them based on Ward's minimum variance criterion
that minimizes 
intra-cluster variance.
Distance between two 
representations is calculated with the squared Euclidean distance. 
The algorithm terminates when the required 
$K$ clusters (aka encoded concepts)  are formed, where $K$ is a hyperparameter. 
Each encoded concept represents a latent relationship between the words present in the cluster. Appendix \ref{sec:appendix:clustering} presents the algorithm.



\subsection{Alignment}
\label{sec:alignment}

Now we define the alignment function between the encoded and human-defined concepts. Consider a human-defined concept as $z$, where a function $z(w)=z$ denotes that $z$ is the human-defined concept of word $w$.  
For example, parts-of-speech is a human-defined concept and each tag such as noun, verb etc. represents a class/label within the concept, 
e.g. $z(sea)=noun$. Similarly, suffix is a human-defined concept 
with various 
suffixes 
representing a class, 
e.g. $z(bigger) = er$. 
A reverse function of z is a one-to-many function that outputs a set of unique words with the given human-defined concept, i.e., $z^{-1}(z)=\{w_1, w_2, \dots, w_J\}$, like $z^{-1}(noun)=\{ sea, tree, \dots\}$, where $J$ is the total number of words with the human-defined concept of $z$. Following this notation, an encoded concept is indicated as $c$, where $c(w)=c$ is a function of applying encoded concept on $w$, and its reverse function outputs a set of unique words with the encoded concept of $c$, i.e., $c^{-1}(c)=\{w_1, w_2, \dots, w_I\}$, where $I$ is the set size. 

To align the encoded concepts 
with the human-defined concepts, we auto-annotate the input data that we used to get the clusters, with the human-defined concepts. 
We call our encoded concept ($c$) to be $\theta$-aligned ($\Lambda_{\theta}$) with a human-defined concept ($z$) as follows:

\begin{equation*}
  \Lambda_{\theta}(z, c)=\left\{
  \begin{array}{@{}ll@{}}
    1, & \text{if}\ \frac{\sum_{w'\in z^{-1}} \sum_{w \in c^{-1}} \delta(w,w')}{J} \geq \theta  \\
    0, & \text{otherwise},
  \end{array}\right.
\end{equation*} 

\noindent where Kronecker function $\delta(w,w')$ is defined as 
\begin{equation*}
  \delta(w,w')=\left\{
  \begin{array}{@{}ll@{}}
    1, & \text{if}\ w=w' \\
    0, & \text{otherwise}
  \end{array}\right.
\end{equation*} 



\noindent We compute $c$ and $\Lambda_{\theta}(z, c)$ for the encoder output from each layer $l$ of a neural network. To compute a network-wise alignment, we simply average $\theta$-agreement over layers.


\section{Experimental Setup}
\label{sec:setup}

\subsection{Dataset}
\label{sec:data}

We used a subset of 
WMT News 2018\footnote{\url{http://data.statmt.org/news-crawl/en/}} (359M tokens) dataset. We randomly selected 250k sentences from the dataset ($\approx$5M tokens) to train our clustering model. 
We discarded words with a frequency of less than 10 and selected maximum 10 occurrences of a word type.\footnote{Our motivation to select a small subset of data and limiting the number of tokens is as follows: clustering a large number of high-dimensional vectors is computationally and memory intensive, for example 200k vectors (of size 768 each) require around 400GB of CPU memory. Applying transformations (e.g., PCA) to reduce dimensionality may result in loss of information and therefore undesirable. We wanted to stay true to the original embeddding space.}
The final dataset consists of 25k word types with 10 contexts per word.

\subsection{Pre-trained Models}
We carried out our analysis on various 12-layered transformer models such as BERT-cased ~\citep[BERT-c, ][]{devlin-etal-2019-bert}, BERT-uncased (BERT-uc), RoBERTa \cite{liu2019roberta}, XLNet \cite{yang2019xlnet} and ALBERT \cite{lan2019albert}. 
We also analyzed multilingual models such as multilingual-bert-cased (mBERT) and XLM-RoBERTa~\citep[XLM-R, ][]{xlm-roberta} where the embedding space is shared across many languages.
This choice of models is motivated from interesting differences in their architectural designs, training data settings (cased vs. un-cased) and multilinguality.

\subsection{Clustering and Alignment}

We extract contextualized representation of words 
by performing a forward pass over the network using the NeuroX toolkit~\cite{neurox-aaai19:demo}. 
We cluster 
representations in every layer 
into $K$ groups. 
To find an optimum value of $K$, we experimented with the 
ELbow~\cite{Thorndike53whobelongs} and 
Silhouette~\cite{silhouetteCluster} methods. However, 
we did not observe reliable results (see Appendix~\ref{sec:appendix:clustering}). 
Therefore, we empirically selected $K=1000$ based on finding a decent balance between many small clusters (over-clustering) and a few large clusters (under-clustering). We found that our results are not sensitive to this parameter and generalize for different cluster settings (See Section~\ref{sec:generalization}).
For the alignment between encoded and human-defined concepts, 
we use $\theta=90\%$ i.e., we consider 
an encoded concept and a human-defined concept to be aligned, if they have at least 90\% match. 



\subsection{Human-defined concepts}
We experiment with the various \textbf{Human-defined concepts}, which we categorize into four groups: 

\begin{itemize}
    \setlength\itemsep{0em}
    \vspace{-1mm}
    \item {\em Lexical Concepts:} Ngrams, Affixes, Casing, First and the Last Word (in a sentence)
    \vspace{-1mm}
    \item {\em Morphology and Semantics:} POS tags \cite{marcus-etal-1993-building} and  SEM tags \cite{abzianidze-EtAl:2017:EACLshort}
    \vspace{-1mm}
    \item {\em Syntactic:} Chunking tags \cite{tjong-kim-sang-buchholz-2000-introduction} and CCG super-tags \citep{hockenmaier2006creating}
    \vspace{-1mm}
    \item {\em Linguistic Ontologies:} WordNet \cite{miller1995wordnet} and LIWC
    \vspace{-1mm}
    ~\citep{pennebaker2001linguistic}
\end{itemize}


At various places in this paper, we also refer to Morphology, Semantics and Syntactic concepts as core-linguistic concepts. We trained BERT-based classifiers using gold-annotated training data and standard splits for each core-linguistic concepts and auto-labelled the selected news dataset using these.\footnote{Please see Appendix \ref{sec:appendix:tagger} for details.} 

\section{Analysis}
\label{sec:analysis}

In this section, we 
analyze the encoded concepts 
by aligning them with the human-defined concepts. 

\begin{table*}
\centering
\footnotesize
\scalebox{1.0}{
\begin{tabular}{l|ccccccc}
\toprule
{} & BERT-c & BERT-uc &  mBERT &  XLM-R & RoBERTa & ALBERT &  XLNet \\
\midrule
Overall alignment                         &  47.2\% &   50.4\% &  66.0\% &  72.4\% &   50.1\% &  51.6\% &  43.6\% \\
\bottomrule
\end{tabular}
}
\vspace{-2mm}
\caption{Coverage of human-defined concepts across all clusters of a given model}
\label{tab:coverage}
\end{table*}




\begin{figure}[t]
    \centering
    \includegraphics[width=\linewidth]{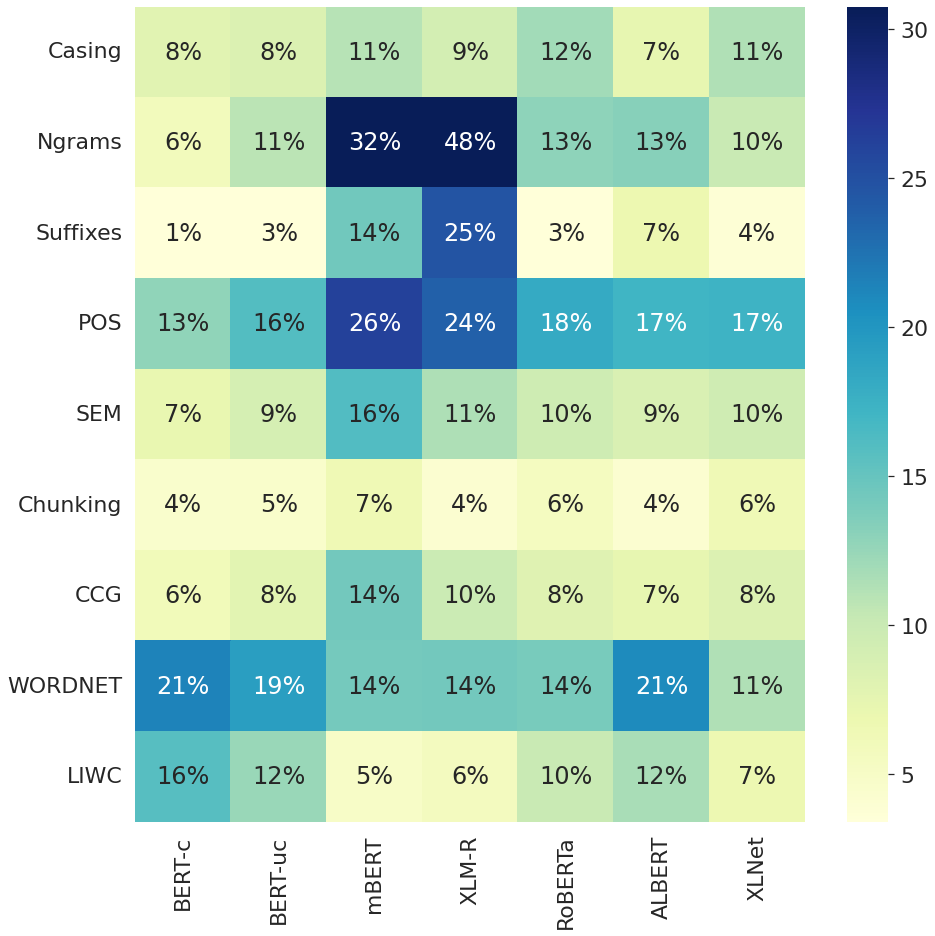}
    \caption{Average Alignment (\%) between encoded concepts and human-defined concepts}
    \label{fig:breakdown}
\end{figure}

\subsection{Overall Alignment}

First we present to what extent the encoded concepts in the entire network align with the human-defined concepts. We compute the overall score as the percentage of the aligned encoded concepts to the human-defined concepts across layers using the function described in Section \ref{sec:alignment}.  
%
We found an overall match of at least 43.6\% in XLNet and at most 72.4\% in XLM-R (See Table \ref{tab:coverage}). Interestingly, the multilingual models (mBERT and XLM-R) found substantially higher match than the monolingual models. The inclusion of multiple languages during training causes the model to learn 
more
linguistic properties. Note that the extent of alignment with the human-defined concept may not necessarily correlate with its overall performance. For example XLNet 
performs 
outperforms BERT on the GLUE tasks, 
but 
aligns less with the human-defined concepts compared to BERT in our results. A similar observation was made by 
\newcite{belinkov-etal-2020-analysis} who 
also found that the translation quality of an NMT model may not correlate with the amount of linguistic knowledge learned in the representation. 
Various factors such as: architectural design, training data, objective function, initialization, etc, play a role in training a pre-trained model. More controlled experiments are needed to understand the relationship of each factor on the performance of the model and on the linguistic learning of the model.

We further investigated per concept\footnote{The first word, last word and prefix concepts showed less less than 1\% alignment with the encoded concepts. We do not present their results in the interest of space.} alignment to understand which human-defined concepts are better represented within the encoded concepts. Figure~\ref{fig:breakdown} presents the results.  
%
%



\paragraph{Lexical Concepts} 
Pre-trained models encode varying amount of lexical concepts such as casing, ngrams and suffixes. 
We found between 7-11\% encoded concepts that align with the casing concept (title case or upper case). We observed that most of these encoded concepts consist of named entities, which were grouped together based on semantics.

\paragraph{Comparing suffixes and ngrams}
While affixes often have linguistic connotation (e.g., the prefix \emph{anti} negates the meaning of the stem and the suffix \emph{ies} is used for pluralization), the ngram units that become part of the vocabulary as an artifact of statistical segmentation (e.g., using BPE \cite{sennrich-etal-2016-neural} or Word-piece \cite{schuster2012japanese}) often lack any linguistic meaning. However, models learn to encode such information. We found a match ranging from 1\% (BERT-cased) up to 25\% (XLM-R) when comparing encoded concepts with the suffix concept. A similar pattern is 
observed in the case of the ngram concept (which is a super-set of the suffix concept) where a staggering 48\% matches were found. 
Figure~\ref{fig:ngram} shows 
an ngram cluster found in layer 2 of BERT-c.\footnote{Appendix \ref{sec:appendix:concept_labels} shows more examples of the ngram, suffix, LIWC and WordNet clusters.}

\begin{figure*}[t]
    \begin{subfigure}[b]{0.24\linewidth}
    \centering
    \includegraphics[width=\linewidth]{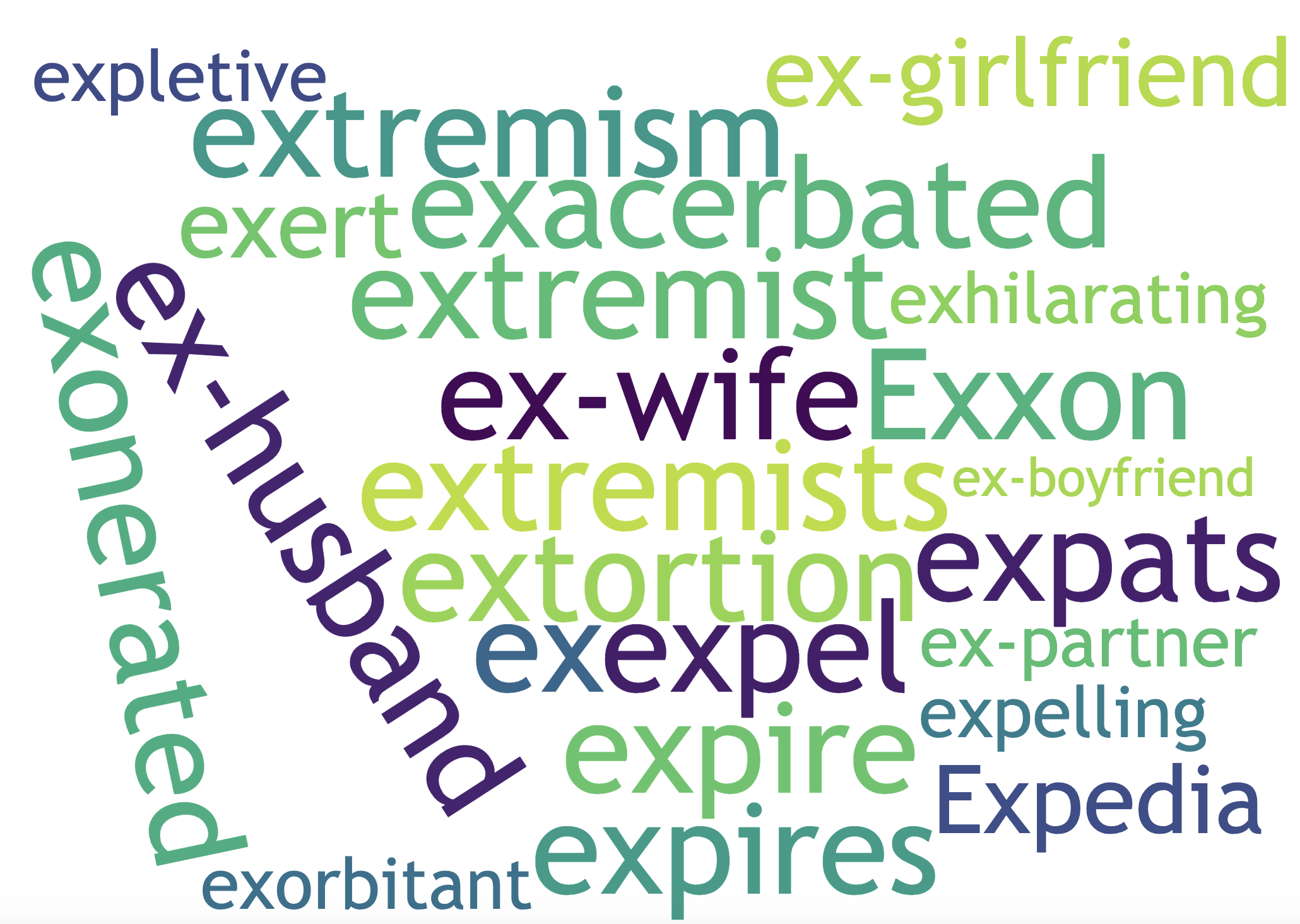}
    \caption{Ngram:ex}
    \label{fig:ngram}
    \end{subfigure}
    \centering
    \begin{subfigure}[b]{0.24\linewidth}
    \centering
    \includegraphics[width=\linewidth]{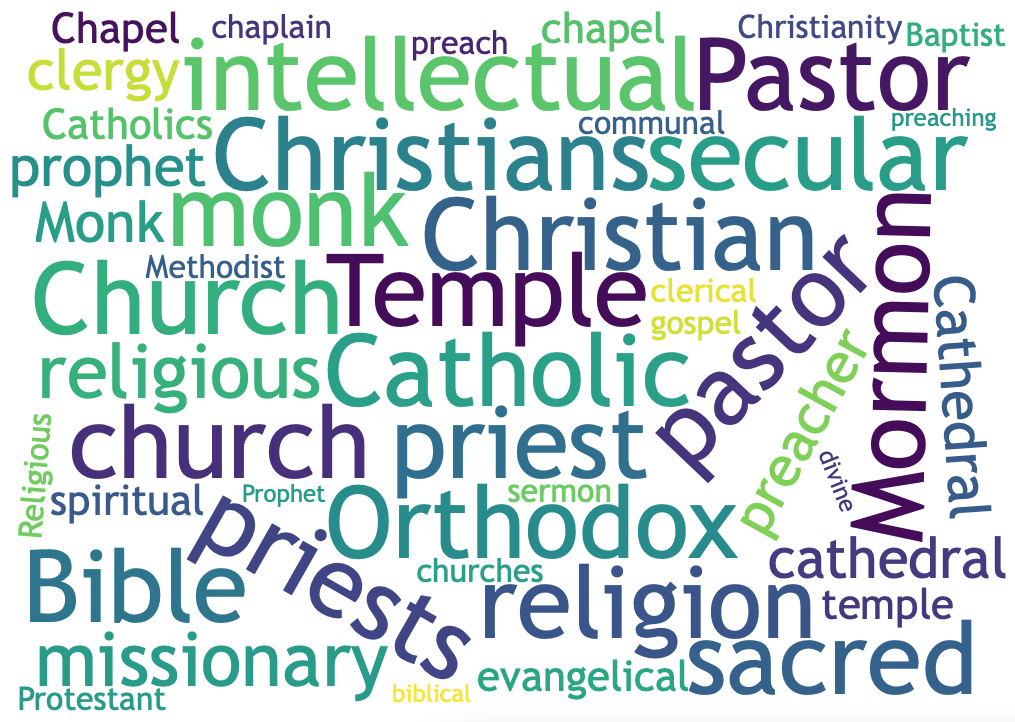}
    \caption{LIWC:religion}
    \label{fig:liwc_religion}
    \end{subfigure}
    \begin{subfigure}[b]{0.24\linewidth}
    \centering
    \includegraphics[width=\linewidth]{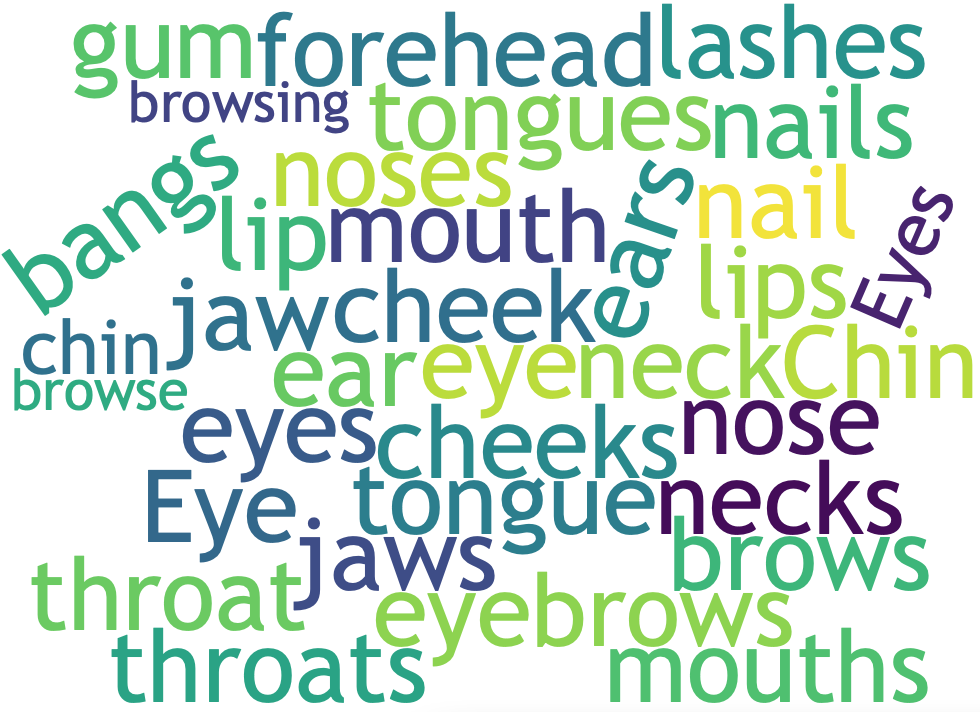}
    \caption{LIWC: Bio}
    \label{fig:liwc_bio}
    \end{subfigure}
    \centering
    \begin{subfigure}[b]{0.24\linewidth}
    \centering
    \includegraphics[width=\linewidth]{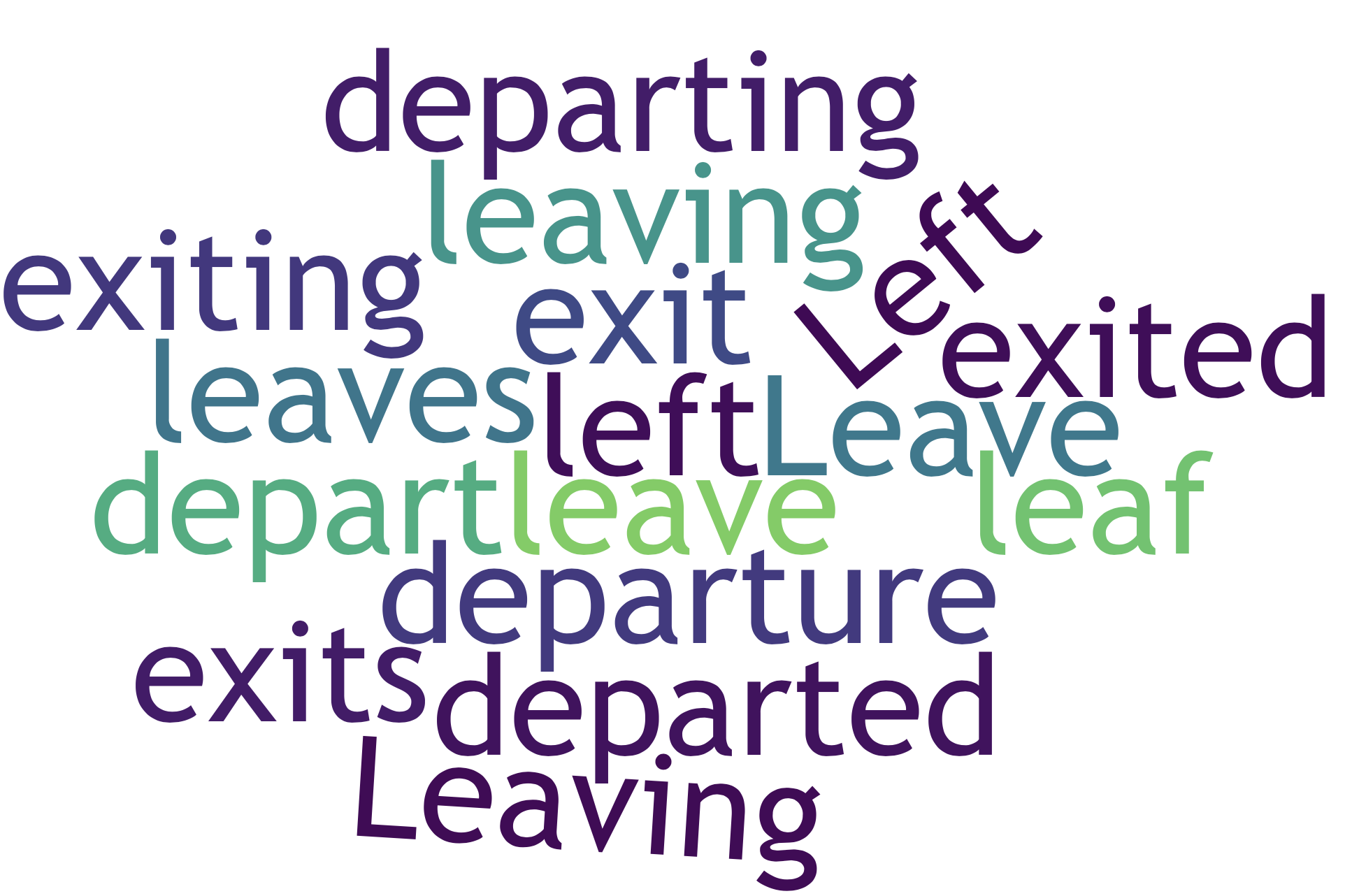}
    \caption{WordNet:Motion}
    \label{fig:wordnet_motion}
    \end{subfigure}     
    \caption{Examples of BERT-c 
    encoded concepts aligned with the human-defined concepts}
    \label{fig:aligned_concept_examples}
\end{figure*}

%
\paragraph{Morphology and Semantics}

We found that the encoded concepts based on word morphology (POS)
consistently showed a higher match across all models in comparison to the other 
abstract concepts, aligning 
a quarter of the encoded concepts in the case of mBERT. The alignment with 
semantic concepts is relatively lower, with at most 16\% match across models. This reflects that while the models learn both linguistic properties, 
morphological ontology is relatively 
preferred compared to the 
semantic hierarchy.
\paragraph{Syntactic}
These concepts capture grammatical orientation of a word, for example 
Chunking:B-NP is a syntactic concept 
describing words in the beginning of a noun phrase. CCG:PP/NP is a concept in CCG super tagging, describing words that takes a noun phrase on the right and outputs a preposition phrase for example ``[in[the US]]''. 
We found relatively fewer matches, a maximum of 7\% and 14\% matching encoded concepts for Chunking and CCG concepts respectively. 
The low matches for syntactic concepts suggest that the  models do not encode the 
same syntactic hierarchy suggested by these human-defined syntactic tasks. 

\paragraph{Linguistic Ontologies} Comparing the encoded concepts with static linguistic ontologies, we found
WordNet concepts 
to 
be the second most 
aligned concept (11-21\%) with the human-defined concepts. 
LIWC also shows a relatively higher 
alignment compared to 
the other human-defined concepts in a few models (e.g., BERT-c). However, 
this observation is not consistent across models and we found a 
range between 5-16\% matches. 
These results present an interesting case where several models prefer the distinction of lexical ontology over abstract linguistic concepts such as morphology. Figure~\ref{fig:aligned_concept_examples} shows examples of encoded concepts aligned with WordNet and LIWC. We see that these concepts are built based on a semantic relationship 
e.g., the clusters in Figure~\ref{fig:liwc_religion},~\ref{fig:liwc_bio} and \ref{fig:wordnet_motion} group words based on religious, 
facial anatomy, 
and specific motion-related vocabulary respectively. 


\paragraph{Comparing Models}

The results of multilingual models (mBERT, XLM-R) are intriguing given that their encoded concepts are dominated by ngram-based concepts and POS concepts, and their relatively
lesser alignment with the linguistic ontologies. On the contrary, several monolingual models (BERT-c, ALBERT) showed a better match with linguistic ontologies specially WordNet.  

The higher number of matches to the ngram (and suffix) concepts in the multilingual models is due to the difference in subword segmentation. 
The subword models in XLM-R and mBERT are optimized for 
multiple languages, resulting in a vocabulary consisting of a large number of small 
ngram units. This causes the multilingual models to aggressively segment the input sequence, compared to the monolingual models\footnote{In our dataset, mBERT has 13\% more words after subword segmentation compared to BERT-c.}  
and resulted in highly dominated ngram-based encoded concepts, especially in the lower layers. This may also explain the relatively lower match that multilingual models exhibit to the linguistic ontologies. We discuss 
this further in the context of layer-wise analysis in Section~\ref{sec:layerwise}. 

Comparing BERT cased vs. uncased, interestingly BERT-uc consistently showed higher matches for the core-linguistic concepts (See Figure \ref{fig:breakdown}). 
We speculate that in the absence of casing information, BERT-uc is forced to learn more linguistic concepts, whereas BERT-c leverages the explicit casing information to capture more semantically motivated concepts based on linguistic ontologies. 

The higher matches in multilingual models in comparison to the  monolingual models, and BERT-uncased in comparison to BERT-cased suggest that the training complexity is one factor that plays a role in a model's ability to learn linguistic nuances. For example, multilingual models need to optimize many languages, which is a harder task compared to learning one language. Similarly, the absence of capitalization in training data makes the learning task relatively harder for BERT-uc compared to BERT-c models, thus resulting in higher matches for BERT-uc. We speculate that the harder the training task, the more language nuances are learned by a model.  \newcite{belinkov-etal-2020-analysis} made a similar observation, where they showed that the linguistic knowledge learned within the encoder-decoder representations in NMT models correlates with complexity of a language-pair involved in 
the task. 

\begin{figure*}[t]
    \centering
    \includegraphics[width=0.95 \linewidth]{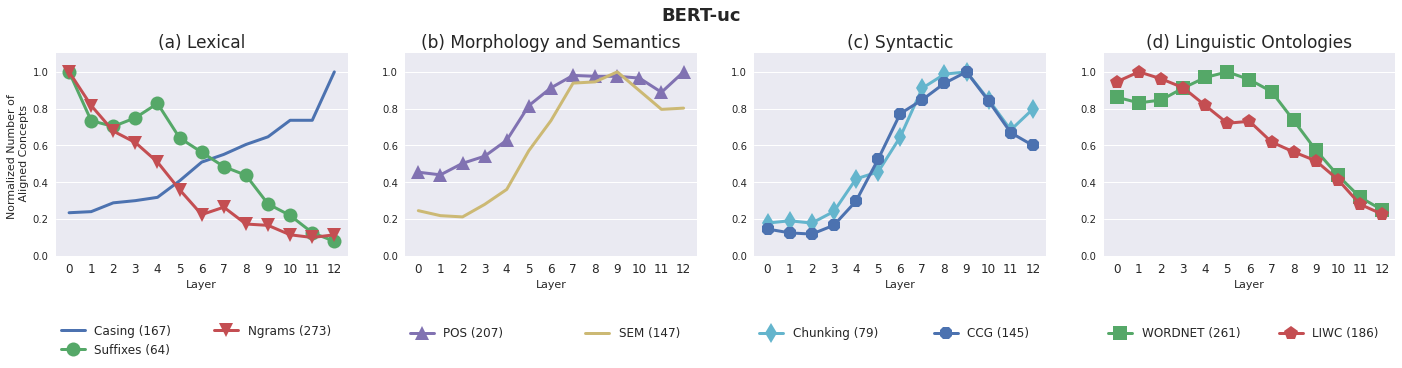}
    \label{fig:bert_uncased}
    \centering
    \includegraphics[width=0.95\linewidth]{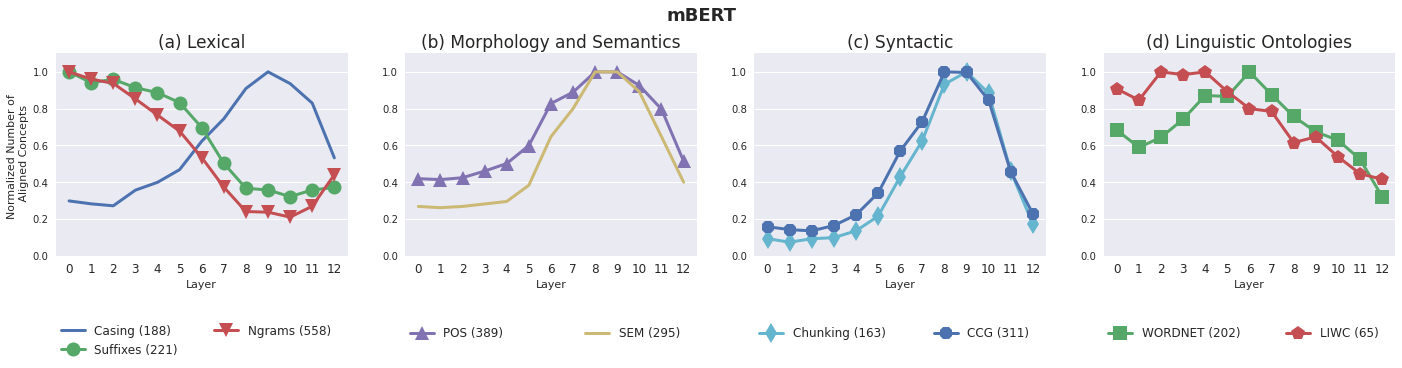}
    \label{fig:mbert}
    \centering
    \includegraphics[width=0.95 \linewidth]{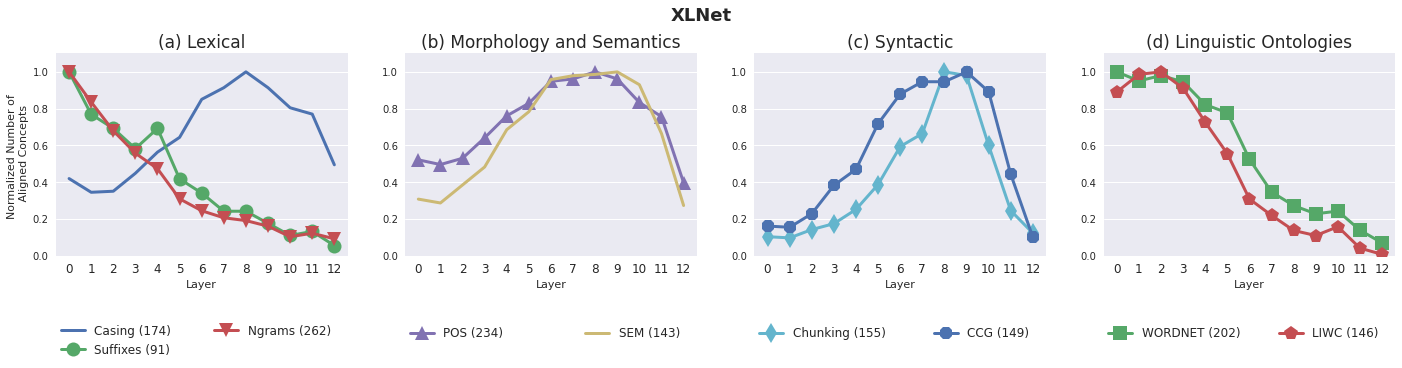}
    \label{fig:xlnet}    
    \centering
    \includegraphics[width=0.95 \linewidth]{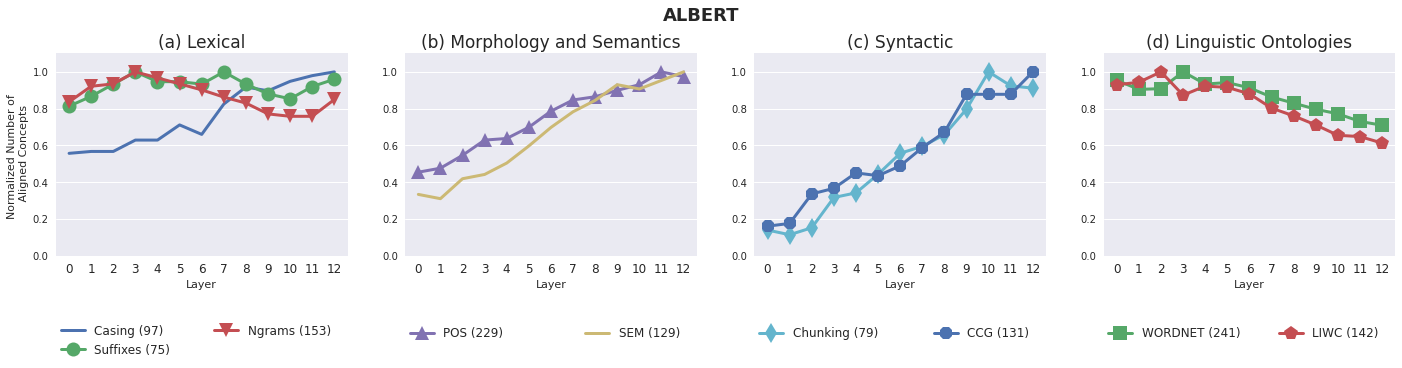}
    \label{fig:albert}
   \caption{Layer-wise concept alignment. Y-axis is the normalized number of aligned concepts. The number within brackets of each human-defined concept, e.g. Casing (166), shows the maximum layer-wise match}
    \label{fig:layerwise}
\end{figure*}

\subsection{Layer-wise Alignment}
\label{sec:layerwise}


Now we study the alignment of human-defined concepts across layers 
to understand
how concepts evolve in the network. 
Figure \ref{fig:layerwise} shows results for selected models.\footnote{See Figure \ref{afig:layerwise} in the Appendix for complete results.} The y-axis is the normalized number of aligned concepts across layers. 
\vspace{-1mm}
\paragraph{Overall Trend}

We observed mostly consistent 
patterns across 
models except for ALBERT, which we will discuss 
later in this section. 
We found that the shallow concepts (such as ngram and suffixes) 
and the linguistic ontologies (LIWC and WORDNET) are better represented in the initial layers and exhibit a downward trend in the higher layers of the network.
On the contrary the core linguistic concepts (POS, Chunking, etc.) are 
better represented in the higher layers (layer 8-10). The last layers do not show 
any consistently dominating human-defined concepts considered in this work. 
We can generalize on these trends and hypothesize on how encoded concepts evolve in the network: the initial layers of the pretrained models, group words based on their lexical and semantic similarities where the former is an artifact of subword segmentation. With the inclusion of context and abstraction in the higher layers, these groups evolve into linguistic manifolds. The encoded concepts in the last layers are influenced by the objective function and 
learn concepts relevant to the task. \newcite{durrani-etal-2021-transfer} also made similar observation when analyzing linguistic concepts in pre-trained models that are fine-tuned towards different GLUE tasks.

\paragraph{Concept-wise Trend} In the following, we discuss different concepts in detail.
As 
we
mentioned 
earlier, the high presence of ngram and suffix concepts in the lower layers is due to 
subword segmentation. At the higher layers, the models start encoding abstract concepts, 
therefore get better alignment with the core linguistic concepts. Casing shows an exception to other lexical concepts and has similar trend to POS and SEM. Upon investigating we observed that the words appearing in these clusters have a hybrid connotation.
For example, more than 98\% of the encoded concepts that match with Casing are named entities, which explains the trend. 
The syntactic concepts 
observe peak in the higher-middle layers and 
a downward trend towards the end. These findings resonate with the earlier work on interpreting neural network representations for BERT. For example \newcite{liu-etal-2019-linguistic} also showed that probes trained with layers 7-8 give the highest accuracy when trained towards predicting the tasks of Chunking and CCG tagging. Although here, we are targeting a slightly different question i.e. how the latent concepts are encoded within the representations and how they evolve from input to output layers of the network. 


We observed a downward trend in linguistic ontologies (WordNet, LIWC) as we go from lower layers to higher layers as opposed to the core linguistic concepts (POS, CCG, etc.). This is because of the context independent nature of these concepts as opposed to the core-linguistic concepts which are annotated based on the context. The embedding layer is non-contextualized, thus shows a high match with linguistic ontologies. With the availability of context in contextualized layers, the encoded concepts evolve into context-aware groups, resulting in higher matches with core-linguistic concepts.

\paragraph{Comparing Models}

While the overall 
trend is consistent among BERT-uc, mBERT and XLNet (and other studied models -- Figure \ref{afig:layerwise} in Appendix), 
the models somewhat differ 
in the last layers: see the large drop in core-linguistic concepts such as POS and Chunking for XLNet and mBERT in comparison to BERT. This 
suggests that 
BERT retains much of the core-linguistic information at the last layers. \newcite{durrani-etal-2020-analyzing} observed a similar pattern in their study, where they showed BERT to retain linguistic information
deeper in the model as opposed to XLNet where it was more localized and predominantly preserved earlier in the network. 
  
While the overall layer-wise trends of multilingual models look similar to some monolingual models (mBERT vs. XLNet in Fig~\ref{fig:layerwise}b,c), the former's absolute layer-wise matches (numbers inside the brackets in Figure~\ref{fig:layerwise} e.g. Casing (166)) are generally substantially higher than the monolingual counterparts. For example, the POS and SEM matches of mBERT are 38.9\% and 30\% respectively which are 18\% and 15\% higher than BERT-uc. On the contrary, the number of matches with linguistic ontologies is often lower for multilingual models (mBERT LIWC alignment of 65 vs. BERT-uc alignment of 186). 
We hypothesize that the variety of training languages in terms of their morphological and syntactic structure has caused the multilingual models to learn more core-linguistic concepts in order to optimize the training task. Although, the knowledge captured within linguistic ontologies is essential, it may not be as critical to the training of the model as the linguistic concepts. 

\textbf{ALBERT} showed a very different trend from the other models. Note that ALBERT shares parameters across layers while the other  
models have separate parameters for every layer. This explains the ALBERT results where we see relatively less variation across layers.
More interestingly, the encoded concepts in the last layers of ALBERT showed presence of all human-defined concepts considered here (see the relatively smaller drop of ALBERT alignment curves in Figure \ref{fig:layerwise}).

\begin{figure*}[t]
    \begin{subfigure}[b]{0.31\linewidth}
    \centering
    \includegraphics[width=\linewidth]{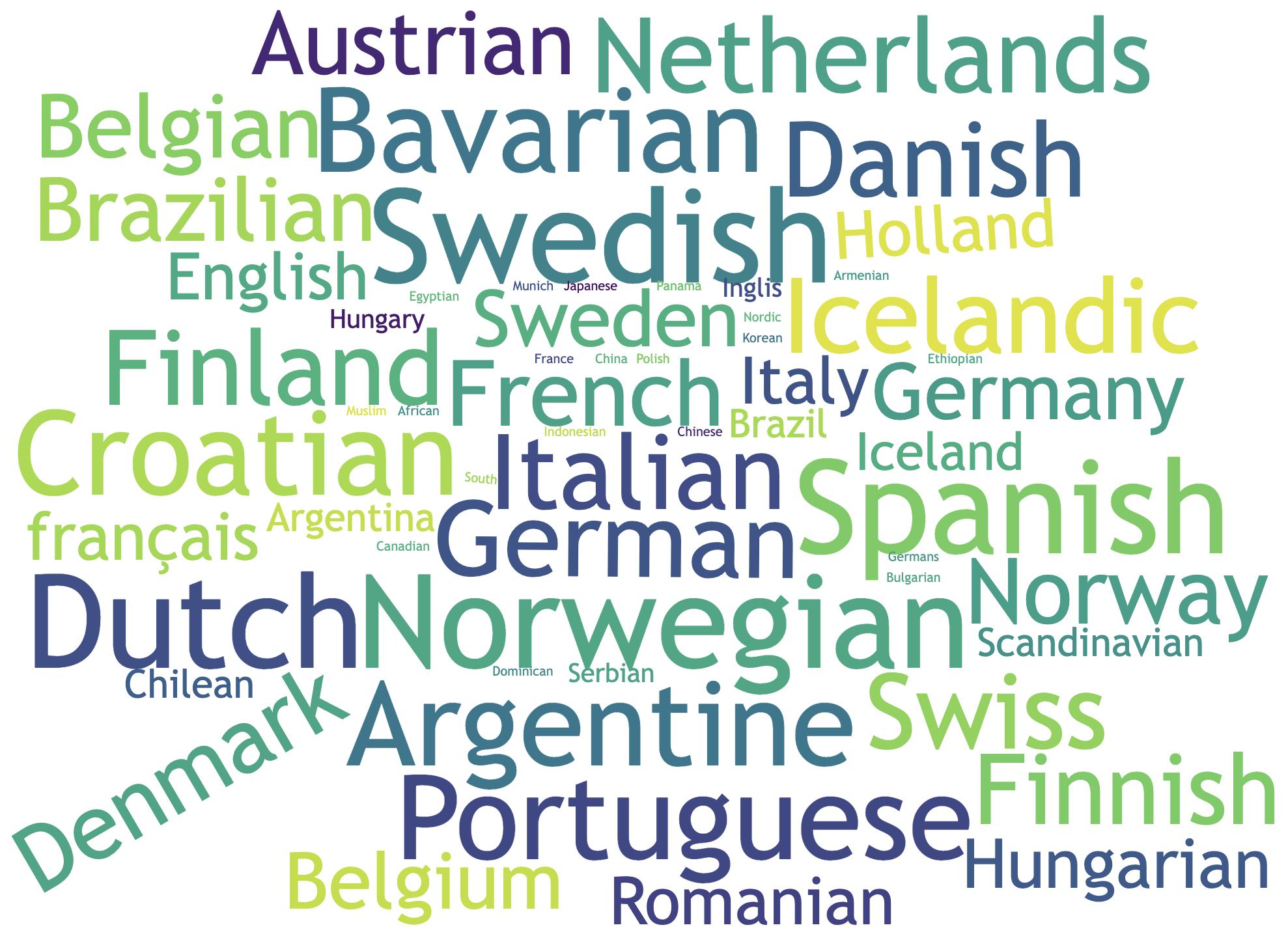}
    \caption{SEM:GPE+POS:JJ}
    \label{fig:jj+nn}
    \end{subfigure}
    \begin{subfigure}[b]{0.31\linewidth}
    \centering
    \includegraphics[width=\linewidth]{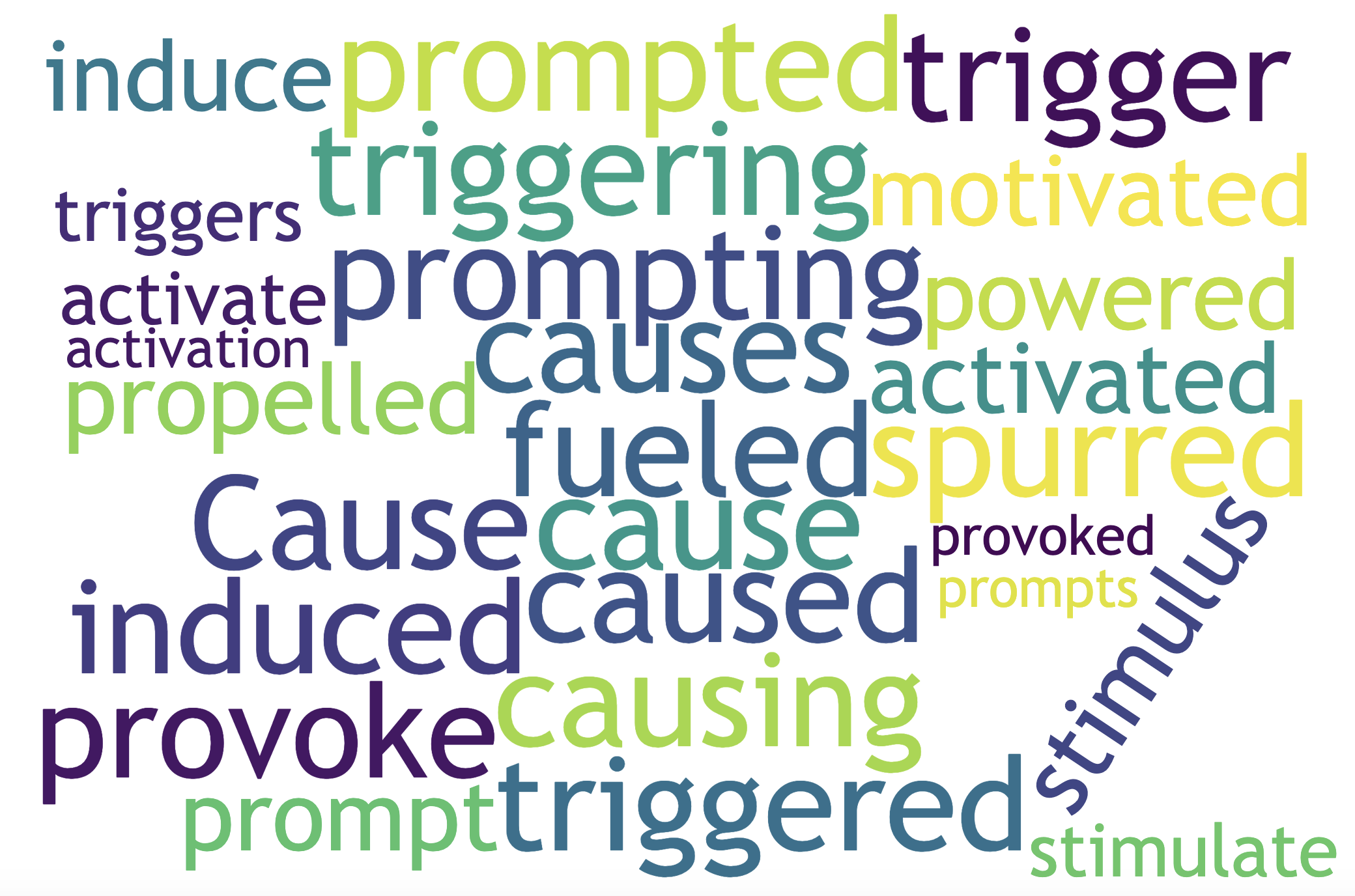}
    \caption{POS:VB*+LIWC:cogmech}
    \label{fig:vbforms}
    \end{subfigure}
    \centering
    \begin{subfigure}[b]{0.31\linewidth}
    \centering
    \includegraphics[width=\linewidth]{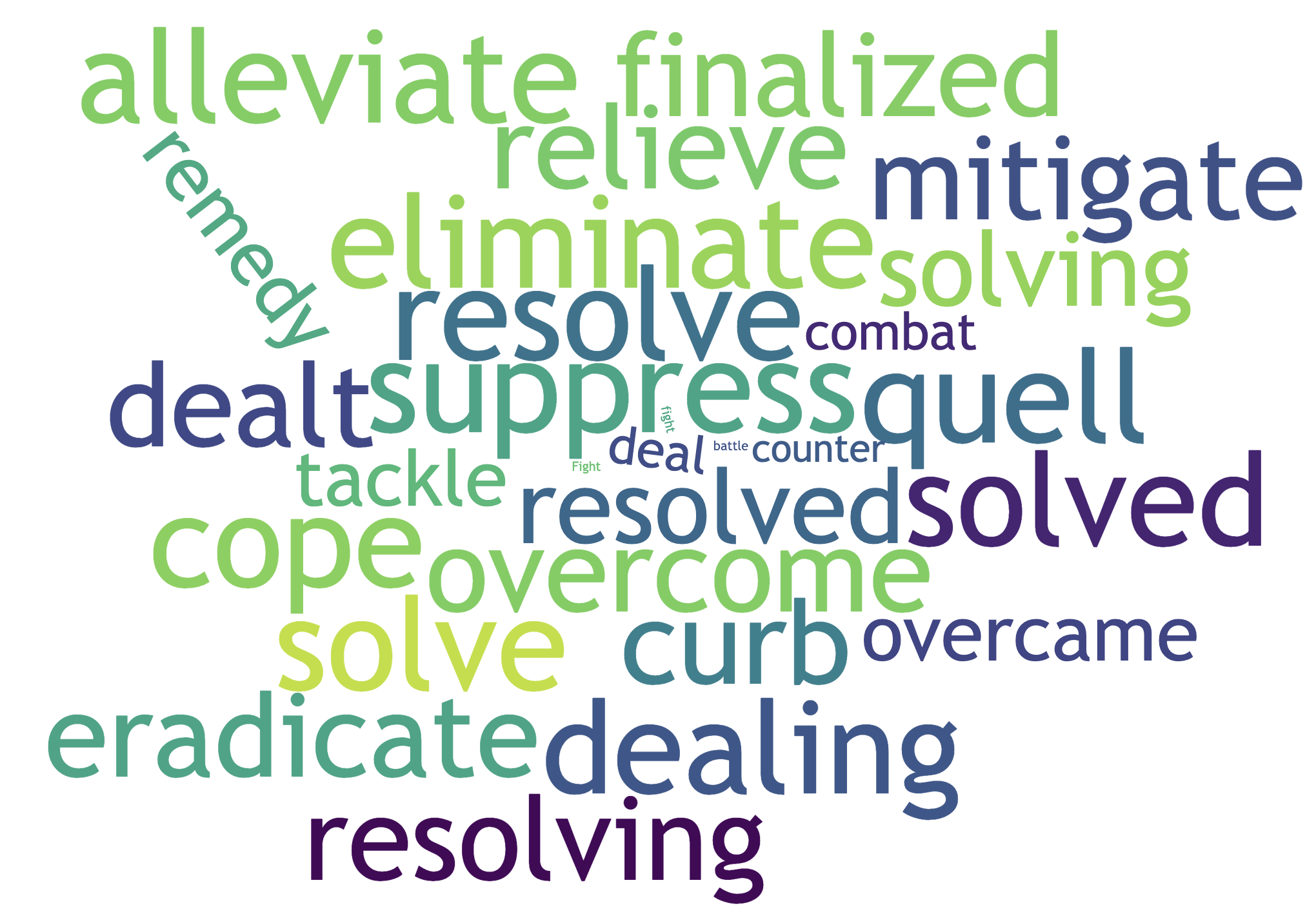}
    \caption{POS:VB*}
    \label{fig:vbforms2}
    \end{subfigure}
    \caption{Examples of unaligned encoded concepts: (a) combination of geopolitical entities and their related adjectives, (b,c) different forms of verb with specific semantics}
    \label{fig:compositional_clusters}
\end{figure*}

\subsection{Unaligned Concepts}
\label{sec:composition}

In Table \ref{tab:coverage} we 
observed that at least 27.6\% (in XLM-R) and up to 56.4\% (in XLNet) encoded concepts did not align with the human-defined concepts. 
\emph{What concepts do these unaligned clusters contain?} In an effort to answer this question, we analyzed these clusters and observed that many of them were \textbf{compositional concepts} that involves more than one fine-grained categories of the human defined concepts. 
Figure~\ref{fig:jj+nn} shows an example of the unaligned concept 
which partly aligns with a semantic category (SEM:geopolitical entity) and a morphological category (POS:adjective). Similarly, Figure \ref{fig:vbforms} is a verbs related to cognitive processes and Figure~\ref{fig:vbforms2} shows an unaligned cluster that is composed of different verb forms (past, present and gerunds). The alignment with multiple human-defined concepts can be used to generate explanations for these unaligned concepts. For example,
Figure~\ref{fig:jj+nn} can be aligned as a mix of geopolitical entities
and adjectives. We also quantitatively verified the number of unaligned encoded concepts that can be explained using composition of different 
concepts (See Appendix \ref{sec:appendix:CC}: Table~\ref{tab:compCoverage}) and found that a majority of the clusters can be explained using a combination of three pre-defined concepts.. 

Moreover, note that encoded concepts are often multifacet i.e., they represent more than one relationship. For example, the encoded concept in Figure~\ref{fig:vbforms2} consists of different forms of verbs but at the same time, these verbs are semantically similar.  
The semantic relationship present here 
is not adequately captured using the human-defined concepts used in this work. These are the \textit{novel concepts} that 
require richer annotations or human-in-the-loop setup to generate adequate explanations. 

\subsection{Generalization of Results}
\label{sec:generalization}
\emph{Do the results generalize over different dataset selection and using different number of clusters?}
We ran experiments using different split of the news dataset for several models, and also performed alignment using different values of $K$, the number of clusters. The results are consistent across the board. Please see Appendix \ref{sec:appendix:method-robustness} for details. 

\section{Conclusion}
\label{sec:conclusion}

We presented \texttt{ConceptX}, a novel framework for analyzing the encoded concepts within deep NLP models. 
Our method uses unsupervised clustering to discover latent concepts within the contextualized representations 
and then aligned these concepts with a suite of human-defined concepts to generate explanations for them. 
Our results illuminate how DNNs structure language information. A few notable findings are: {\em (i)} lower layers capture shallow linguistic concepts, 
{\em (ii)} whereas the abstract linguistic concepts such as morphology and semantics are preserved higher in the network, {\em (iii)} the extent of alignment varies across different models and different human-defined concepts, {\em (iv)} we found that novel explanations and an improved coverage of concepts can be achieved via compositionality. 


\bibliography{bib/anthology,bib/custom}
\bibliographystyle{acl_natbib}

\newpage
\clearpage
\section*{Appendix}
\label{sec:appendix}
\appendix

\section{Human-defined concept labels}
\label{sec:appendix:concept_labels}

\subsection{Lexical Concepts:}
Ngrams, Affixes, Casing, First and the Last Word.

\subsection{Morphology and Semantics:}
\paragraph{POS tags:}
We used the Penn Treebank POS tags discussed in \cite{marcus-etal-1993-building}, which consists of 36 POS tags and 12 other tags (i.e., punctuation and currency symbols). In Table \ref{tab:penn_treebank_pos_tags}, we provide POS tags and their description. 

\begin{table}[!tbh]
\centering
\scalebox{0.75}{
\setlength{\tabcolsep}{2.0pt}
\begin{tabular}{@{}lll@{}}
\toprule
\textbf{\#} & \textbf{Tag} & \textbf{Description} \\ \midrule
1 & CC & Coordinating conjunction \\
2 & CD & Cardinal number \\
3 & DT & Determiner \\
4 & EX & Existential there \\
5 & FW & Foreign word \\
6 & IN & Preposition or subordinating conjunction \\
7 & JJ & Adjective \\
8 & JJR & Adjective, comparative \\
9 & JJS & Adjective, superlative \\
10 & LS & List item marker \\
11 & MD & Modal \\
12 & NN & Noun, singular or mass \\
13 & NNS & Noun, plural \\
14 & NNP & Proper noun, singular \\
15 & NNPS & Proper noun, plural \\
16 & PDT & Predeterminer \\
17 & POS & Possessive ending \\
18 & PRP & Personal pronoun \\
19 & PRP\$ & Possessive pronoun \\
20 & RB & Adverb \\
21 & RBR & Adverb, comparative \\
22 & RBS & Adverb, superlative \\
23 & RP & Particle \\
24 & SYM & Symbol \\
25 & TO & to \\
26 & UH & Interjection \\
27 & VB & Verb, base form \\
28 & VBD & Verb, past tense \\
29 & VBG & Verb, gerund or present participle \\
30 & VBN & Verb, past participle \\
31 & VBP & Verb, non-3rd person singular present \\
32 & VBZ & Verb, 3rd person singular present \\
33 & WDT & Wh-determiner \\
34 & WP & Wh-pronoun \\
35 & WP\$ & Possessive wh-pronoun \\
36 & WRB & Wh-adverb \\
37 & \# & Pound sign \\
38 & \$ & Dollar sign \\
39 & . & Sentence-final punctuation \\
40 & , & Comma \\
41 & : & Colon, semi-colon \\
42 & ( & Left bracket character \\
43 & ) & Right bracket character \\
44 & " & Straight double quote \\
45 & ' & Left open single quote \\
46 & " & Left open double quote \\
47 & ' & Right close single quote \\
48 & " & Right close double quote \\ \bottomrule
\end{tabular}%
}
\caption{Penn Treebank POS tags.}
\label{tab:penn_treebank_pos_tags}
\end{table}
 
\paragraph{SEM tags:} \cite{abzianidze-EtAl:2017:EACLshort} consists of 73 sem-tags grouped into 13 meta-tags. In Table \ref{tab:sem-tags}, we provide a detailed information of the tagset, and in Table \ref{atab:semCoarse}, we provide fine and coarse tags mapping. 

\begin{table*}[!tbh]
\centering
\scalebox{0.75}{
\setlength{\tabcolsep}{2.0pt}
\begin{tabular}{@{}llll@{}}
\toprule
\textbf{ANA   (anaphoric)} &  & \multicolumn{2}{l}{\textbf{MOD   (modality)}} \\ \midrule
PRO & anaphoric \& deictic pronouns: he, she, I, him & NOT & negation: not, no, neither, without \\
DEF & definite: the, loIT, derDE & NEC & necessity: must, should, have to \\
HAS & possessive pronoun: my, her & POS & possibility: might, could, perhaps, alleged, can \\
REF & reflexive \& reciprocal pron.: herself, each other & \multicolumn{2}{l}{\textbf{DSC (discourse)}} \\
EMP & emphasizing pronouns: himself & SUB & subordinate relations: that, while, because \\
\textbf{ACT (speech act)} &  & COO & coordinate relations: so, \{,\}, \{;\}, and \\
GRE & greeting \& parting: hi, bye & APP & appositional relations: \{,\}, which, \{(\}, — \\
ITJ & interjections, exclamations: alas, ah & BUT & contrast: but, yet \\
HES & hesitation: err & \multicolumn{2}{l}{\textbf{NAM (named entity)}} \\
QUE & interrogative: who, which, ? & PER & person: Axl Rose, Sherlock Holmes \\
\textbf{ATT (attribute)} &  & GPE & geo-political entity: Paris, Japan \\
QUC\* & concrete quantity: two, six million, twice & GPO\* & geo-political origin: Parisian, French \\
QUV\* & vague quantity: millions, many, enough & GEO & geographical location: Alps, Nile \\
COL\* & colour: red, crimson, light blue, chestnut brown & ORG & organization: IKEA, EU \\
IST & intersective: open, vegetarian, quickly & ART & artifact: iOS 7 \\
SST & subsective: skillful surgeon, tall kid & HAP & happening: Eurovision 2017 \\
PRI & privative: former, fake & UOM & unit of measurement: meter, \$, \%, degree Celsius \\
DEG\* & degree: 2 meters tall, 20 years old & CTC\* & contact information: 112, info@mail.com \\
INT & intensifier: very, much, too, rather & URL & URL: \url{http://pmb.let.rug.nl} \\
REL & relation: in, on, 's, of, after & LIT\* & literal use of names: his name is John \\
SCO & score: 3-0, grade A & NTH\* & other names: table 1a, equation (1) \\
\multicolumn{2}{l}{\textbf{COM   (comparative)}} & \multicolumn{2}{l}{\textbf{EVE (events)}} \\
EQU & equative: as tall as John, whales are mammals & EXS & untensed simple: to walk, is eaten, destruction \\
MOR & comparative positive: better, more & ENS & present simple: we walk, he walks \\
LES & comparative negative: less, worse & EPS & past simple: ate, went \\
TOP & superlative positive: most, mostly & EXG & untensed progressive: is running \\
BOT & superlative negative: worst, least & EXT & untensed perfect: has eaten \\
ORD & ordinal: 1st, 3rd, third & \multicolumn{2}{l}{\textbf{TNS (tense \& aspect)}} \\
\multicolumn{2}{l}{\textbf{UNE   (unnamed entity)}} & NOW & present tense: is skiing, do ski, has skied, now \\
CON & concept: dog, person & PST & past tense: was baked, had gone, did go \\
ROL & role: student, brother, prof., victim & FUT & future tense: will, shall \\
GRP\* & group: John \{,\} Mary and Sam gathered, a group of people & PRG\* & progressive: has been being treated, aan hetNL \\
\multicolumn{2}{l}{\textbf{DXS (deixis)}} & PFT\* & perfect: has been going/done \\
DXP\* & place deixis: here, this, above & \multicolumn{2}{l}{\textbf{TIM (temporal entity)}} \\
DXT\* & temporal deixis: just, later, tomorrow & DAT\* & full date: 27.04.2017, 27/04/17 \\
DXD\* & discourse deixis: latter, former, above & DOM & day of month: 27th December \\
\multicolumn{2}{l}{\textbf{LOG (logical)}} & YOC & year of century: 2017 \\
ALT & alternative \& repetitions: another, different, again & DOW & day of week: Thursday \\
XCL & exclusive: only, just & MOY & month of year: April \\
NIL & empty semantics: \{.\}, to, of & DEC & decade: 80s, 1990s \\
DIS & disjunction \& exist. quantif.: a, some, any, or & CLO & clocktime: 8:45 pm, 10 o'clock, noon \\
IMP & implication: if, when, unless &  &  \\
AND & \multicolumn{2}{l}{conjunction \& univ. quantif.:   every, and, who, any} &  \\ \bottomrule
\end{tabular}%
}
\caption{Semantic tags.}
\label{tab:sem-tags}
\end{table*}

\subsection{Syntactic:}
\paragraph{Chunking tags:}
For Chunking we used the tagset discussed in ~\cite{tjong-kim-sang-buchholz-2000-introduction}, which consists of 11 tags as follows: 
NP (Noun phrase), 
VP (Verb phrase), 
PP (Prepositional phrase), 
ADVP (Adverb phrase), 
SBAR (Subordinate phrase), 
ADJP (Adjective phrase), 
PRT (Particles), 
CONJP (Conjunction), 
INTJ (Interjection), 
LST (List marker), 
UCP (Unlike coordinate phrase). For the annotation, chunks are represented using IOB format, which results in 22 tags in the dataset as reported in Table \ref{tab:dataStats}. 

\paragraph{CCG super-tags}~\newcite{hockenmaier2006creating} developed, CCGbank, a dataset with Combinatory Categorial Grammar (CCG) derivations and dependency structures from the Penn Treebank. CCG is a lexicalized grammar formalism, which is expressive and efficiently parseable. It consists of 1272 tags.  

\subsection{Linguistic Ontologies:}
\paragraph{WordNet:}~\cite{miller1995wordnet} consists of 26 lexicographic senses for nouns, 2 for adjectives, and 1 for adverbs. Each of them represent a supersense and a hierarchy can be formed from hypernym to hyponym. 


\paragraph{LIWC:} Over the past few decades, Pennebaker et al. \cite{pennebaker2001linguistic} have designed psycholinguistic concepts using high frequency words. These word categories are mostly used to study gender, age, personality, and health to estimate the correlation between these attributes and word usage. It is a knowledge-based system where words are mapped different high level concepts.

\begin{table}[h]									
\centering					
\scalebox{0.76}{
\setlength{\tabcolsep}{2.5pt}
    \begin{tabular}{l|cccc|c}									
    \toprule									
Task    & Train & Dev & Test & Tags & F1\\		
\midrule
    POS & 36557 & 1802 & 1963 & 48 & 96.69\\
    SEM & 36928 & 5301 & 10600 & 73 & 96.22 \\
    Chunking &  8881 &  1843 &  2011 & 22 & 96.91 \\
    CCG &  39101 & 1908 & 2404 & 1272 & 94.90\\
    \bottomrule
    \end{tabular}
    }
    \caption{Data statistics (number of sentences) on training, development and test sets using in the experiments and the number of tags to be predicted}

\label{tab:dataStats}						
\end{table}

\section{BERT-based Sequence Tagger}
\label{sec:appendix:tagger}

We trained a BERT-based sequence tagger to auto-annotate our training data. We used standard splits for training, development and test data for the 4 linguistic tasks (POS, SEM, Chunking and CCG super tagging) that we used to carry out our analysis on. The splits to preprocess the data are available through git repository\footnote{\url{https://github.com/nelson-liu/contextual-repr-analysis}} released with \newcite{liu-etal-2019-linguistic}. See Table \ref{tab:dataStats} for statistics and classifier accuracy.

\clearpage
\section{Clustering details}
\label{sec:appendix:clustering}

Algorithm~\ref{alg1} assigns each word to a separate cluster and then iteratively combines them based on Ward's minimum variance criterion
that minimizes 
intra-cluster variance.
Distance between two vector representations is calculated with the squared Euclidean distance. 

	\begin{algorithm}[!tbh]
		\caption{Clustering Procedure} 
		\label{alg1} 
		\textbf{Input}: $\overrightarrow{y}^l$: word representation of words\\
		\textbf{Parameter}: $K$: the total number of clusters
		\begin{algorithmic}[1] 
			\FOR{each word $w_i$}
			\STATE assign $w_i$ to cluster $c_i$
			\ENDFOR
			\WHILE{number of clusters $\neq K$}
			\FOR{each cluster pair $c_i$,$c_{i^{\prime}}$}
			\STATE $d_{i,i^{\prime}}$ = inner-cluster difference of combined cluster $c_i$ and $c_{i^{\prime}}$
			\ENDFOR
			\STATE $c_j$,$c_{j^{\prime}}$ = cluster pair with minimum value of $d$
			\STATE merge clusters $c_j$ and $c_{j^{\prime}}$
			\ENDWHILE
		\end{algorithmic}
	\end{algorithm}


\subsection{Selection of the number of Clusters}
The Elbow curve did not show any optimum clustering point, with the increase in number of clusters the distortion score kept decreasing, resulting in over-clustering (a large number of clusters consisted of less than 5 words). The over-clustering resulted in high but wrong alignment scores e.g. consider a two word cluster having words “good” and “great”. The cluster will have a successful match with “adjective” since more than 90\% of the words in the cluster are adjectives. In this way, a lot of small clusters will have a successful match with many human-defined concepts and the resulting alignment scores will be high. On the other hand, Silhouette resulted in under-clustering, giving the best score at number of clusters = 10. 
We handled this empirically by trying several values for the number of clusters i.e., 200 to 1600 with step size 200. We selected 1000 to find a good balance with over and under clustering. We understand that this may not be the best optimal point. We presented the results of 600 and 1000 clusters to show that our findings are not sensitive to the number of clusters parameter.

\section{Coarse vs. Fine-grained Categories}
\label{asec:coarsefine}
\subsection{Coarse vs. Fine-grained Categories}
Our analysis of compositional concepts showed that several fine-grained concepts could be combined to explain an unaligned concept. For example, by combining verb categories of POS to one coarse verb category, we can align the encoded concept present in Figure \ref{fig:vbforms2}. To probe this more formally, we collapsed POS and SEM fine-grained concepts into coarser categories (27 POS tags and 15 SEM tags).
We then recomputed the alignment with the encoded concepts.
For most of the models, the alignment doubled compared to the fine-grained categorizes with at least 39\% and at most 53\% percent match for POS. This reflects that in several cases, models learn the coarse language hierarchy.
%
We further questioned \textit{how many encoded concepts can be explained using coarse human-defined concepts.} 
Compared to Table \ref{tab:coverage}, the matches increased by at most 17 points in the case of BERT-uc. The XLM-R showed the highest matching percentage of 81\%. The higher alignment suggests that most of the encoded concepts learned by pre-trained models can be explained using human-defined concepts. (See Appendix~\ref{asec:coarsefine} for detailed results).

\subsection{Corase POS and SEM labels}
\label{asec:coarse}

Tables \ref{atab:semCoarse} and \ref{atab:posCoarse} present results for our mapping of fine-grained SEM and POS tags into coarser categories. 

\begin{table}[h]
\scalebox{0.80}{
\begin{tabular}{l|l}
\toprule
Coarse & Fine-grained \\ 
\midrule
ACT & QUE \\
ANA & DEF, DST, EMP, HAS, PRO, REF \\
ATT & INT, IST, QUA, REL, SCO \\
COM & COM, LES, MOR, TOP \\
DSC & APP, BUT, COO, SUB \\
DXS & PRX \\
EVE & EXG, EXS, EXT, EXV \\
LOG & ALT, AND, DIS, EXC, EXN, IMP, NIL, RLI \\
MOD & NEC, NOT, POS \\
NAM & ART, GPE, HAP, LOC, NAT, ORG, PER, UOM \\
TIM & DEC, DOM, DOW, MOY, TIM, YOC \\
TNS & EFS, ENG, ENS, ENT, EPG, EPS, \\ 
       & EPT, ETG, ETV, FUT, NOW, PST \\
UNE & CON, ROL \\
UNK & UNK \\
\bottomrule
\end{tabular}
}
\caption{SEM: Coarse to Fine-grained mapping}
\label{atab:semCoarse}
\end{table}

\begin{table}[!tbh]
\scalebox{0.80}{
\begin{tabular}{l|l}
\toprule
Coarse & Fine-grained \\ 
\midrule
Adjective   & JJ, JJR, JJS   \\
Adverb      & RB, RBS, WRB, RBR \\
Conjunction & CC \\
Determiner  & DT, WDT      \\
Noun        & NN, NNS, NNP, NNPS \\
Number      & CD \\
Preposition & IN, TO \\
Pronoun     & PRP, PRP\$, WP, WP\$  \\
Verb        & VB, VBN, VBZ, VBG, VBP, VBD \\
No Changes    & \$, -LRB-, \#, FW, -RRB-, LS, POS, "", EX\\ 
              & SYM, ,, :, RP, ., PDT, MD, UH, \\
              
\bottomrule
\end{tabular}
}
\caption{POS: Coarse to Fine-grained mapping}
\label{atab:posCoarse}
\end{table}

\subsection{Results}
\label{asec:coarseresults}

Table~\ref{atab:coarsevsfine} presents the alignment results of using coarse POS and SEM concepts. We observed that the alignment doubles in most of the cases which reflects that in several cases, models learn the coarse language hierarchy. However, they do not strictly adhere to fine-grained categories existed in human-defined concepts. We further extend the alignment of coarse POS and SEM categories to the overall alignment with the human-defined concepts. Table \ref{atab:overall_coarse} presents the results. We see a match of up to 81\% in the case of XLM-R. The high alignment suggests that many of the encoded concepts can be explained using coarse human-defined concepts.

\begin{table}[h]
\footnotesize
\begin{tabular}{l|cc|cc}
\toprule
{} & \multicolumn{2}{c}{POS} & \multicolumn{2}{c}{SEM} \\
{} & Fine & Coarse & Fine & Coarse \\
\midrule
BERT-cased   &  13\% &    42\% &   7\% &    15\% \\
BERT-uncased &  16\% &    43\% &   9\% &    18\% \\
mBERT        &  26\% &    53\% &  16\% &    26\% \\
XLM-RoBERTa  &  24\% &    47\% &  11\% &    21\% \\
RoBERTa      &  18\% &    43\% &  10\% &    20\% \\
ALBERT       &  17\% &    42\% &   9\% &    17\% \\
XLNet        &  17\% &    39\% &  10\% &    18\% \\
\bottomrule
\end{tabular}
\caption{Alignment of fine-grained human defined concepts compared to coarse categories}
\label{atab:coarsevsfine}
\end{table}

\begin{table} []
\centering
\footnotesize
\begin{tabular}{l|cccc}
\toprule
{} & \textbf{BERT-c} & \textbf{BERT-uc} &  \textbf{mBERT} &  \textbf{XLM-R}  \\

Overall   &  61.5\% &   63.6\% &  77.7\% &  81.0\% \\
alignment & \textbf{RoBERTa} & \textbf{ALBERT} &  \textbf{XLNet} \\
&   62.9\% &  64.0\% &  55.3\% \\
\bottomrule
\end{tabular}
\caption{Coverage of human-defined concepts using coarse POS and SEM labels across all clusters from a given model}
\label{atab:overall_coarse}
\end{table}

\begin{figure*}[!tbh]
    \begin{subfigure}[b]{0.30\linewidth}
    \centering
    \includegraphics[width=\linewidth]{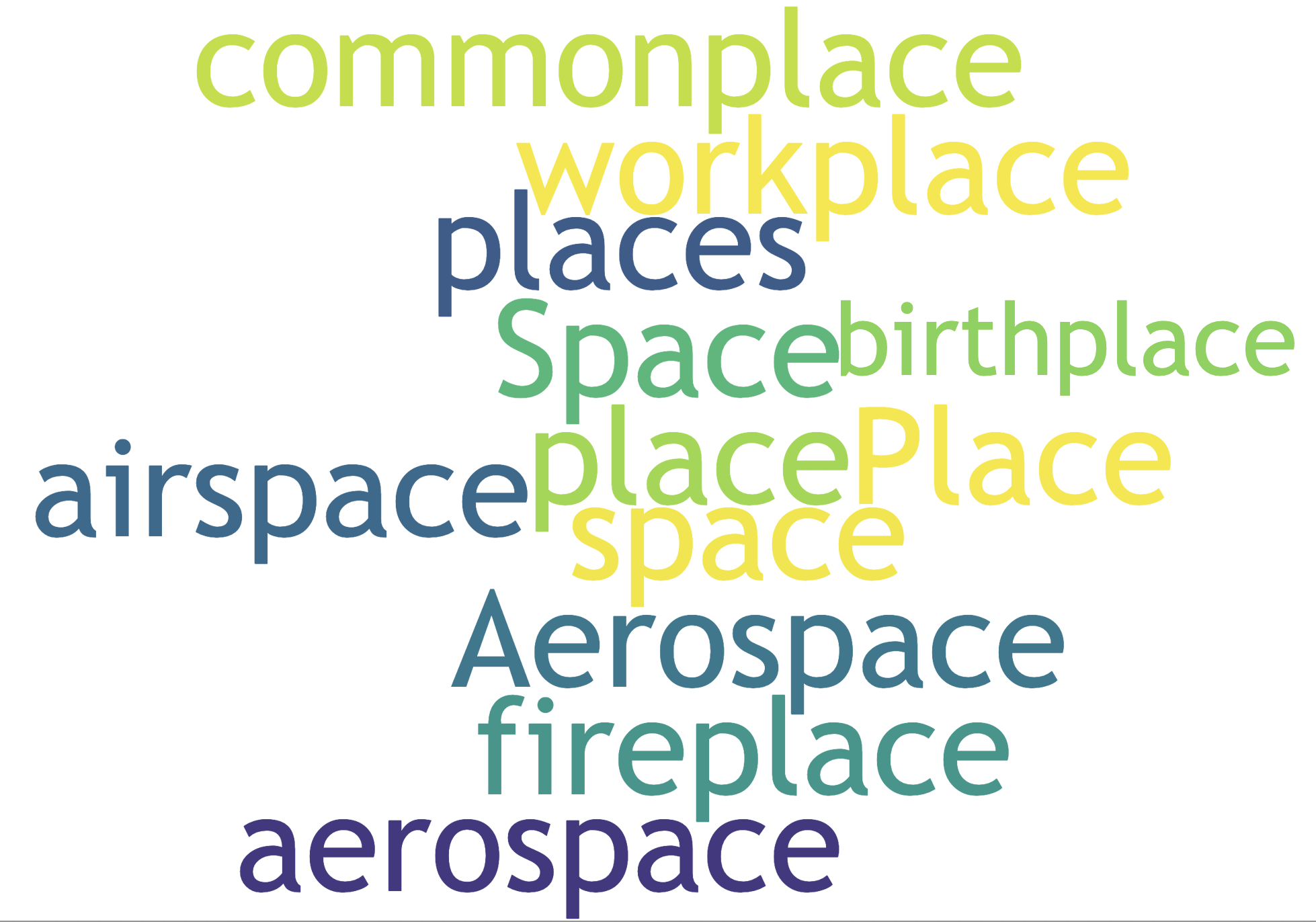}
    \caption{}
    \label{fig:ngram}
    \end{subfigure}
    \centering
    \begin{subfigure}[b]{0.30\linewidth}
    \centering
    \includegraphics[width=\linewidth]{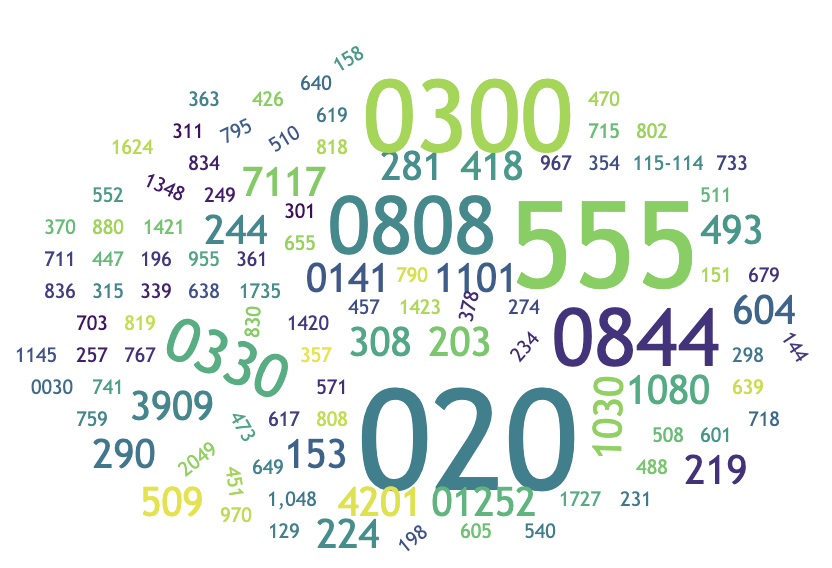}
    \caption{}
    \label{fig:cd}
    \end{subfigure}    
    \centering
    \begin{subfigure}[b]{0.30\linewidth}
    \centering
    \includegraphics[width=\linewidth]{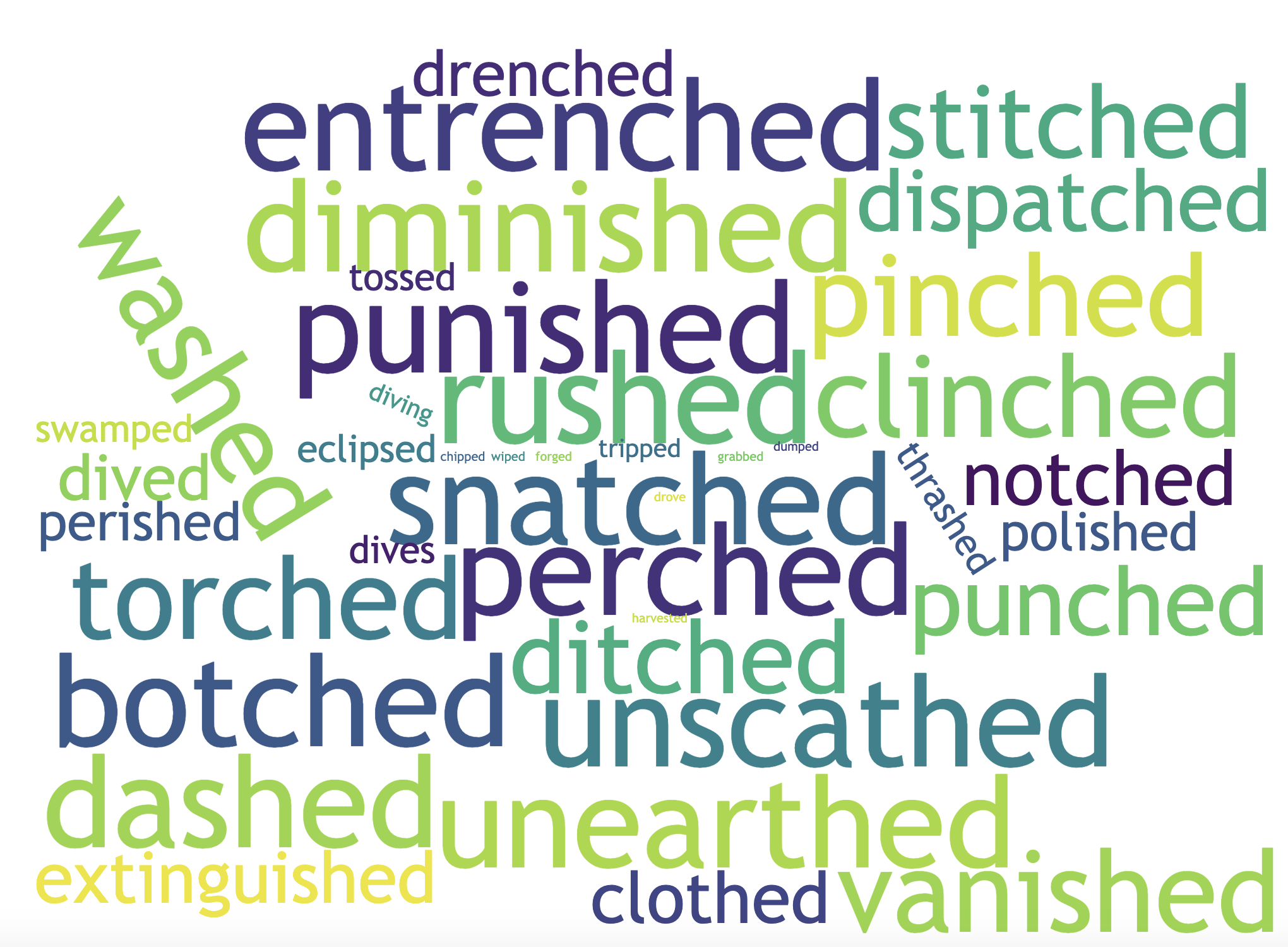}
    \caption{}
    \label{fig:artifact}
    \end{subfigure}
    \caption{Example clusters: (a) ngram:ace, (b) POS:CD, (c) Chunking:B-VP + Suffix:ed}
    \label{fig:clusters-sup}
\end{figure*}

\begin{figure*}[!tbh]
    \begin{subfigure}[b]{0.30\linewidth}
    \centering
    \includegraphics[width=\linewidth]{figures/BERT-c-L1-232.png}
    \caption{}
    \label{fig:cause}
    \end{subfigure}
    \centering
    \begin{subfigure}[b]{0.30\linewidth}
    \centering
    \includegraphics[width=\linewidth]{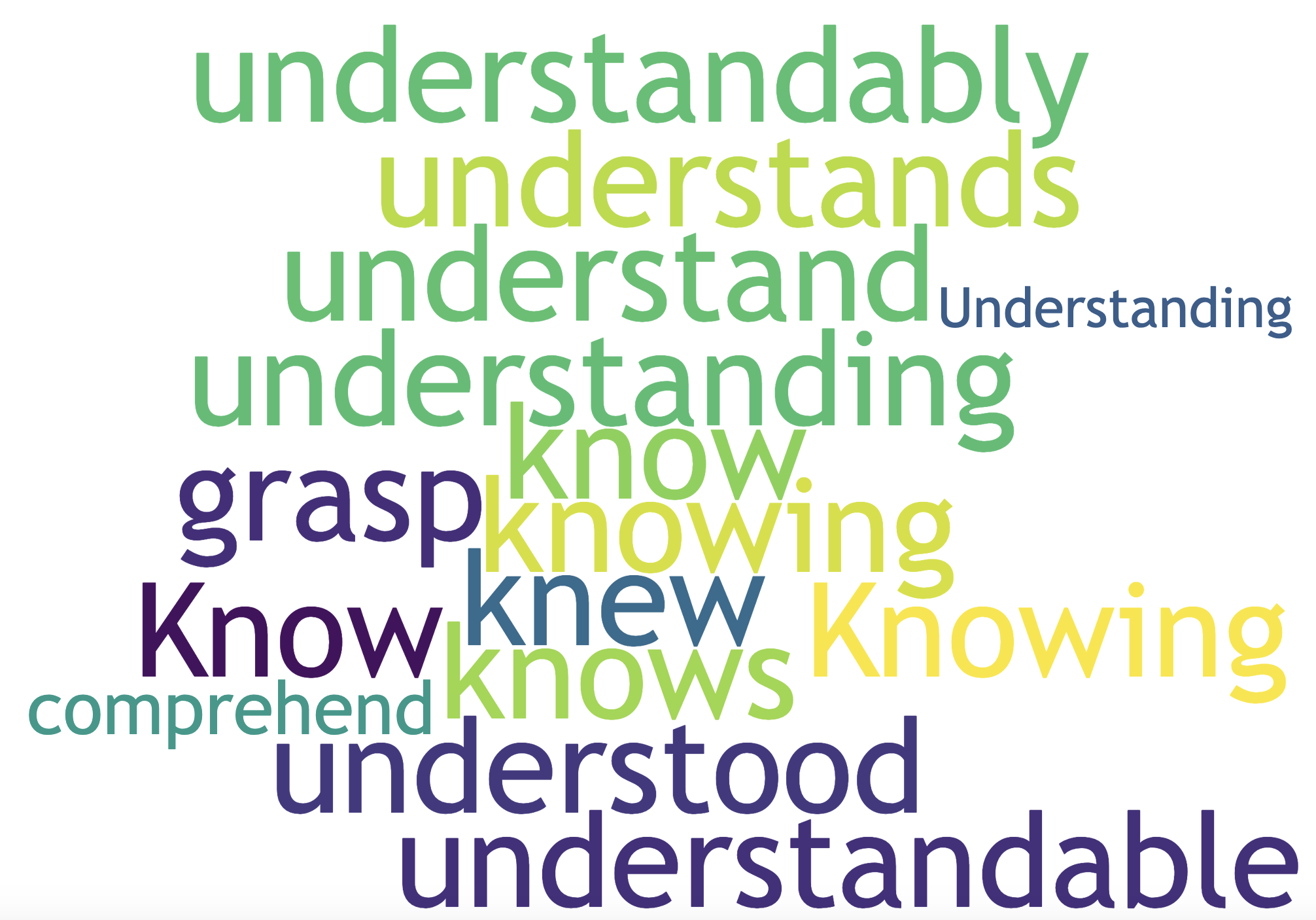}
    \caption{}
    \label{fig:cognition}
    \end{subfigure}    
    \centering
    \begin{subfigure}[b]{0.30\linewidth}
    \centering
    \includegraphics[width=\linewidth]{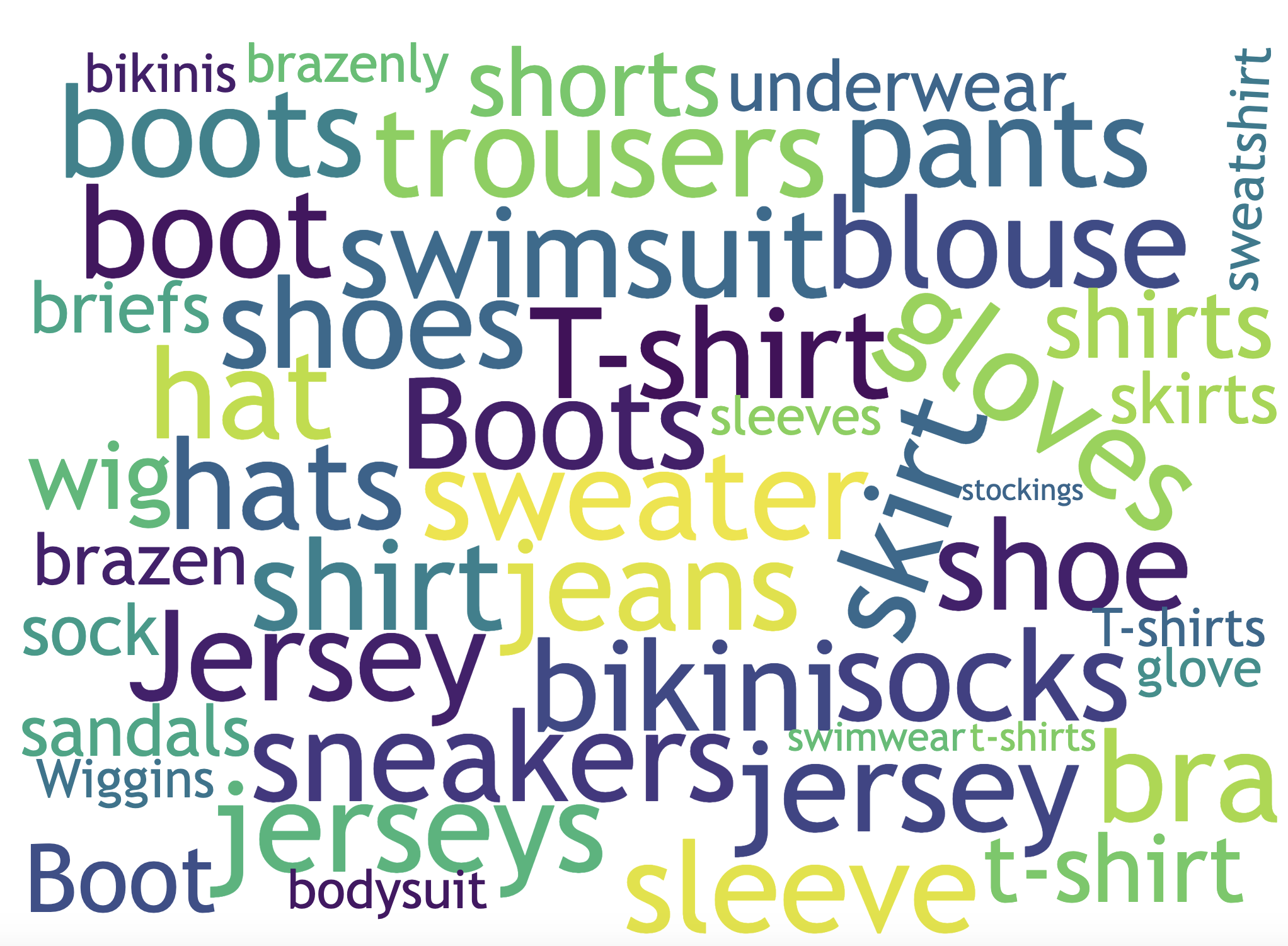}
    \caption{}
    \label{fig:artifact}
    \end{subfigure}
    \caption{Example clusters: (a) LIWC:cause, (b) WORDNET:verb.cognition, (c) WORDNET:noun.artifact}
    \label{fig:clusters-sup-ontology}
\end{figure*}

\section{Compositional Coverage}
\label{sec:appendix:CC}

Table \ref{tab:compCoverage} shows the amount of coverage we obtain when aligning with the morphological concepts when allowing 90\% of the words in the cluster to be from $N$ concepts.

\begin{table*}[!tbh]									
\centering
\footnotesize
    \begin{tabular}{c|ccccccc}									
    \toprule									
Concepts     & BERT-c & BERT-uc & mBERT & XLM-R & RoBERTa & ALBERT & XLNet \\		
\midrule
    1 & 13\% & 16\% & 26\% & 24\% & 18\% & 17\% & 17\%\\
    2 & 11\% & 12\% & 20\% & 23\% & 13\% & 13\% & 12\%\\
    3 & 14\% & 13\% & 14\% & 18\% & 11\% & 15\% & 9\%\\
    4 & 6\% & 6\% & 4\% & 4\% & 5\% & 5\% & 3\%\\
    5 & 2\% & 1\% & 1\% & 1\% & 2\% & 1\% & 1\%\\
    6 & 1\% & 0\% & 0\% & 0\% & 1\% & 1\% & 0\%\\
    \bottomrule
    \end{tabular}
    \caption{Percentage of alignment when an encoded concept is composed of $N$ morphological concepts. As can be seen, most concepts are composed of either 1, 2 or 3 morphological concepts, showing that several concepts learned by these models are indeed compositional in nature. }

\label{tab:compCoverage}
\end{table*}

\section{Robustness of Methodology across Datasets and Settings}
\label{sec:appendix:method-robustness}

\begin{figure*}[!tbh]
    \centering
    \includegraphics[width=0.9\linewidth]{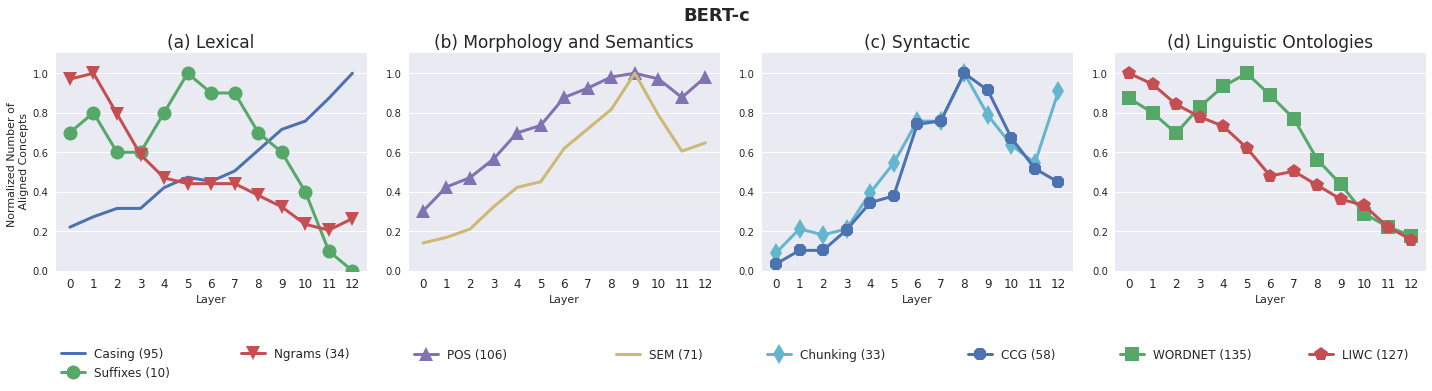}
    \label{fig:600-bert-c}
    \centering
    \includegraphics[width=0.9\linewidth]{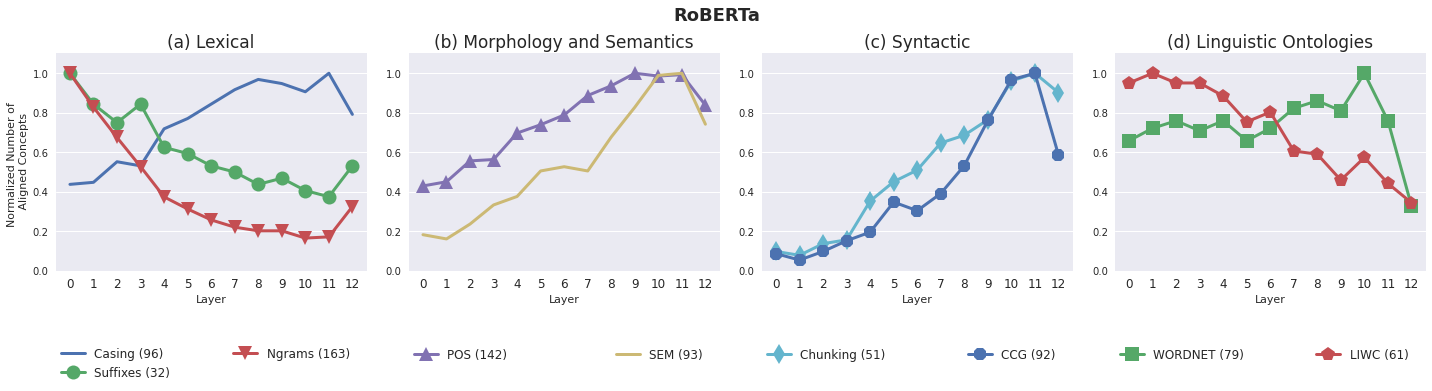}
    \label{fig:600-roberta}
    \includegraphics[width=0.9 \linewidth]{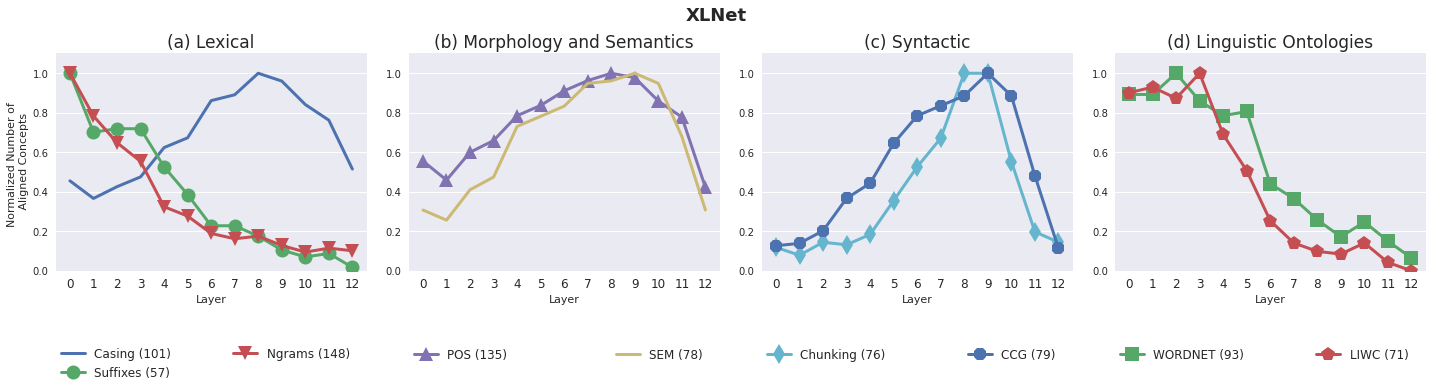}
    \label{fig:600-xlnet} 
        \centering
    \includegraphics[width=0.9\linewidth]{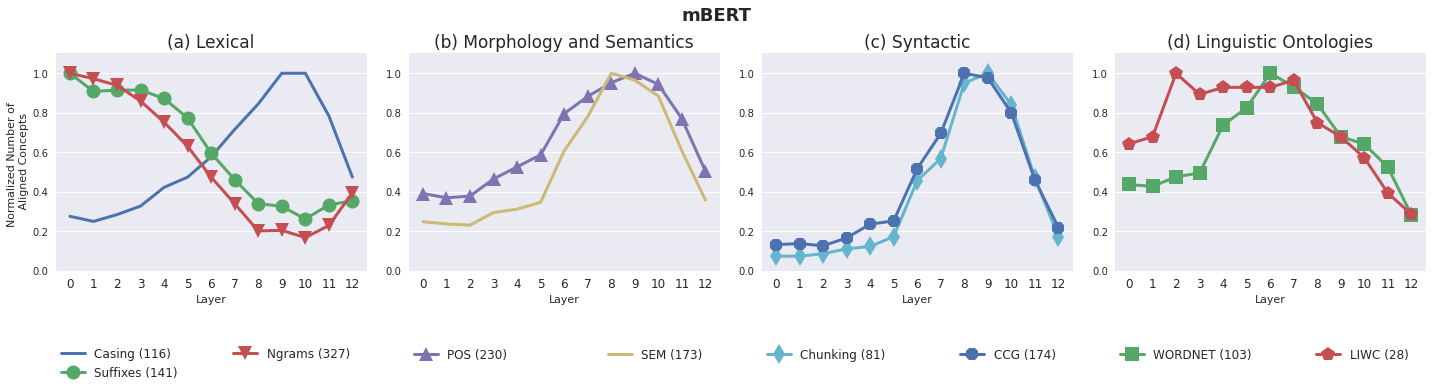}
    \label{fig:600-mbert}
   \caption{Layer-wise results using 600 clusters.}
    \label{afig:layerwise-600}
\end{figure*}

Figure \ref{afig:layerwise-600} shows the layer-wise patterns using 600 clusters instead of 1000 as used in the main paper. We observe that 
the overall trends largely remain the same. 

\begin{figure*}[!tbh]
    \centering
    \includegraphics[width=0.9\linewidth]{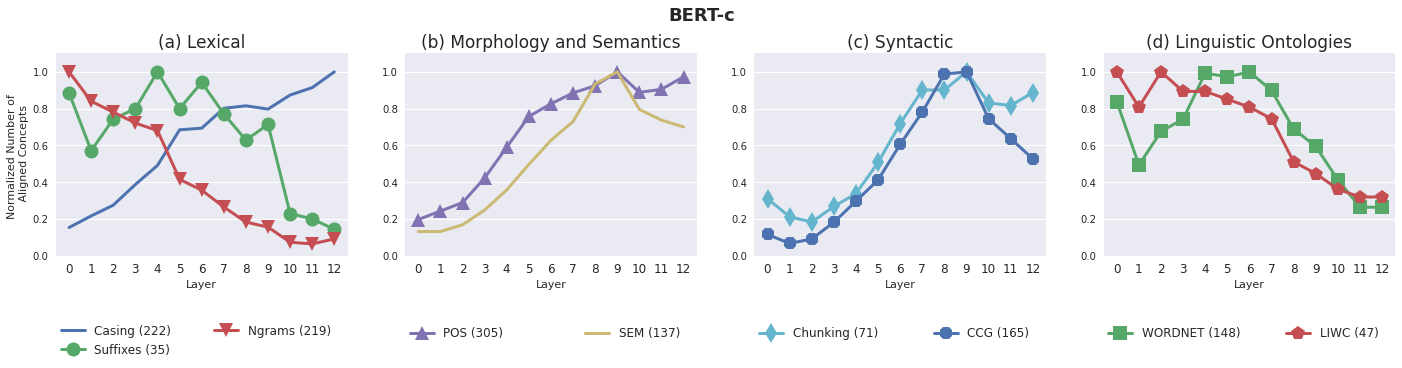}
    \label{fig:diff-data-bert-c}
    \includegraphics[width=0.9 \linewidth]{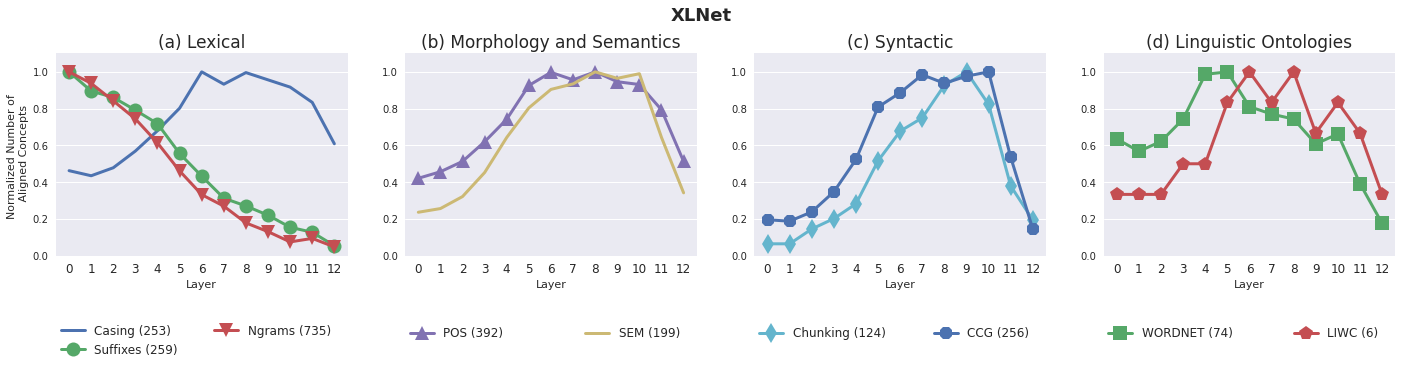}
    \label{fig:diff-data-xlnet}    
   \caption{Layer-wise results on a separately sampled dataset.}
    \label{afig:layerwise-diff-data}
\end{figure*}

To further demonstrate the robustness of our method with respect to dataset, we sub-sampled another dataset from the News corpus with a different vocabulary by selecting words that appear between 2 to 10 times in the corpus. Note that the selection of vocabulary is due to the memory and computation limitations. Figure \ref{afig:layerwise-diff-data} shows the results using this selection of data. Compared to Figure~\ref{fig:layerwise}, we can see that the overall patterns are largely similar and confirms the robustness of our findings. The slight difference in the patterns of WordNet and LIWC are due to the large selection of proper nouns in the second set of the data.

\section{Layer-wise results}
\label{asec:layerwise}

\begin{figure*}[!tbh]
    \centering
    \includegraphics[width=0.82\linewidth]{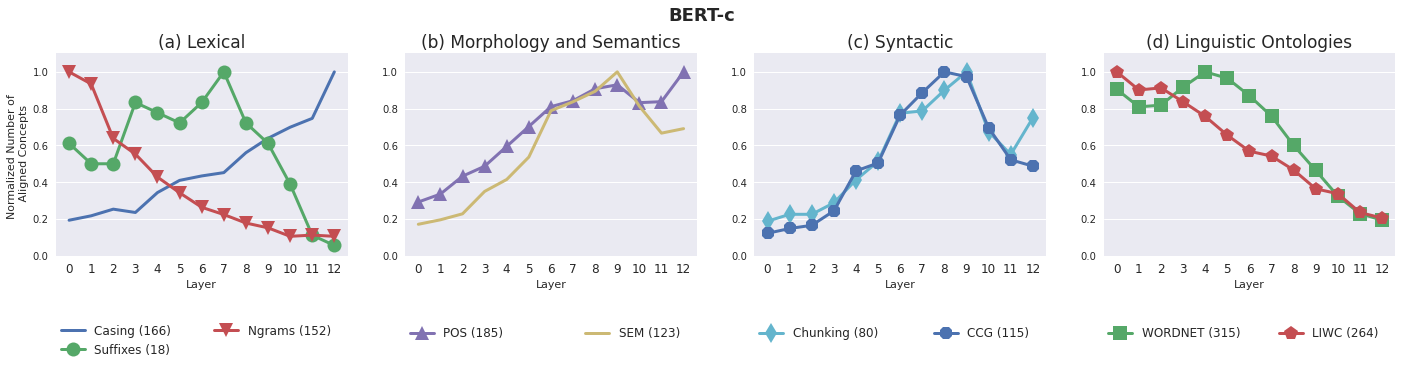}
    \label{fig:bert-c}
    \centering
    \includegraphics[width=0.82 \linewidth]{figures/layerwise-bert-base-uncased_news_first_250k_closed_class.png}
    \label{fig:bert_uncased}
    \centering
    \includegraphics[width=0.82\linewidth]{figures/layerwise-bert-base-multilingual-cased_news_first_250k_closed_class.png}
    \label{fig:mbert}
    \centering
    \includegraphics[width=0.82\linewidth]{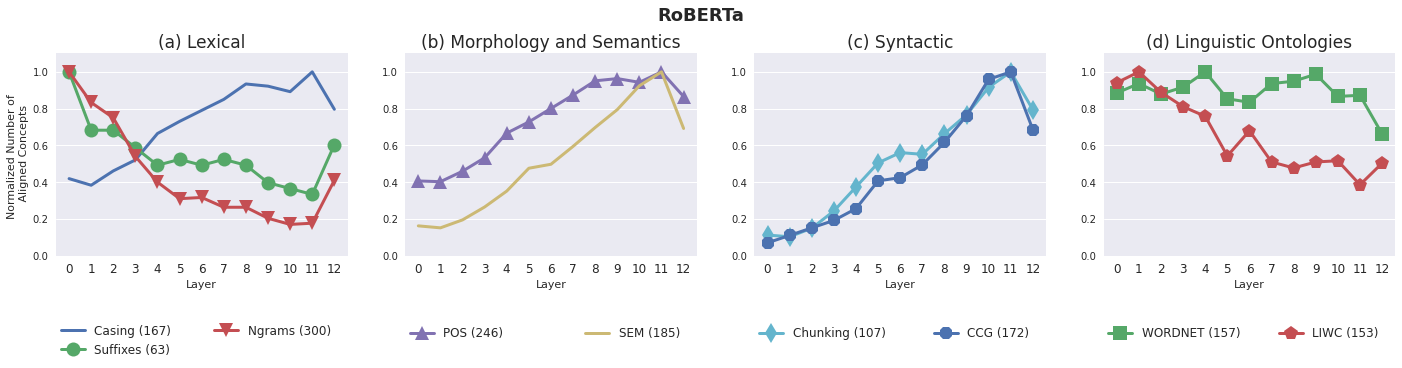}
    \label{fig:roberta}
    \centering
    \includegraphics[width=0.82\linewidth]{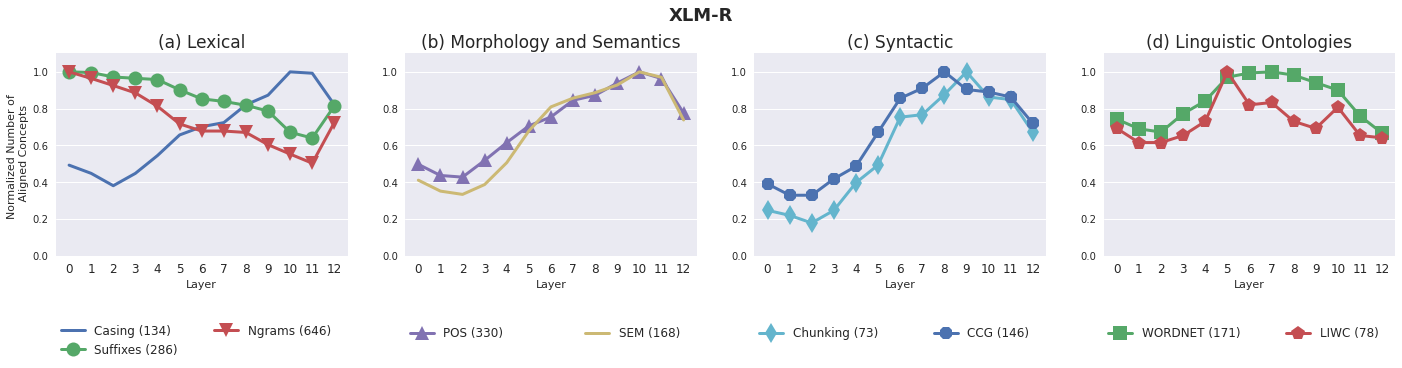}
    \label{fig:xlm-roberta}
    \centering
    \includegraphics[width=0.82 \linewidth]{figures/layerwise-albert-base-v1_news_first_250k_closed_class.png}
    \label{fig:albert}

    \centering
    \includegraphics[width=0.82\linewidth]{figures/layerwise-xlnet-base-cased_news_first_250k_closed_class.png}
    \label{fig:xlnet}    
   \caption{Layer-wise results.}
    \label{afig:layerwise}
\end{figure*}

Figure \ref{afig:layerwise} present layer-wise results for all the understudied models. 

\end{document}